\documentclass[final, times, numbers, sort&compress]{elsarticle}
\usepackage{lineno,hyperref}
\modulolinenumbers[5]
\journal{Journal of \LaTeX\ Templates}

\usepackage{amsthm}
\usepackage{graphicx}
\usepackage{amsmath}
\usepackage{mathtools}
\usepackage{enumitem}

\usepackage[ruled,linesnumbered,resetcount]{algorithm2e}
\usepackage{algpseudocode}
\usepackage{booktabs}
\usepackage[table]{xcolor}
\usepackage{pdflscape}
\usepackage{longtable}
\usepackage{multirow}
\usepackage{geometry}
\usepackage{threeparttablex}
\usepackage{graphicx}
\usepackage[compatibility=false]{caption}
\usepackage[labelformat=simple]{subcaption}
\usepackage[numbers]{natbib}
\usepackage{url}
\usepackage{rotating}

\usepackage{hyperref}
\usepackage{lipsum}
\usepackage{import}
\usepackage[acronym]{glossaries}
\makeglossaries

\usepackage{graphicx}
\usepackage{amsmath}
\usepackage{mathtools}
\usepackage[ruled,linesnumbered,resetcount]{algorithm2e}
\usepackage{algpseudocode}
\usepackage{float}
\usepackage{pdfpages}
\usepackage{import}
\usepackage{multicol}
\usepackage{lineno}
\usepackage{tabularx}
\usepackage{setspace}
\usepackage{soul}
\hypersetup{
	bookmarks=false,
	colorlinks=true,       
	linkcolor=red,          
	citecolor=green,        
	filecolor=magenta,      
	urlcolor=cyan         
}
\newacronym{ga}{GA}{Genetic Algorithm}
\newacronym{ea}{EA}{Evolutionary Algorithm}
\newacronym{mfo}{MFO}{Multifactorial Optimization}
\newacronym{mfea}{MFEA}{Multifactorial Evolutionary Algorithm}
\newacronym{cluspt}{CluSPT}{Clustered Shortest-Path Tree Problem}
\newacronym{sto}{STO}{single-tasking of Genetic Algorithm}
\newacronym{clumrct}{CluMRCT}{Minimum Routing Cost Clustered Tree Problem}
\newacronym{clutsp}{CluTSP}{Clustered Traveling Salesman Problem}
\newacronym{ftsp}{FTSP}{Family Traveling Salesman Problem}
\newacronym{gtsp}{GTSP}{Generalized Traveling Salesman Problem}

\newacronym{tsp}{TSP}{Traveling Salesman Problem}
\newacronym{steinertp}{STP}{Steiner Tree Problem}
\newacronym{clusteinertp}{CluSteinerTP}{Clustered Steiner Tree Problem}
\newacronym{uss}{USS}{Unified Search Space}
\newacronym{de}{DE}{Differential Evolution Algorithm}
\newacronym{pso}{PSO}{Particles Warm Optimization Algorithm}
\newacronym{rmp}{rmp}{Probability of Random Mating}

\newacronym{cesa}{CESA}{}
\newacronym{iim}{IIM}{New Individual Initialization Method}
\newacronym{ncx}{NCX}{New Crossover Operator}
\newacronym{nmo}{NMO}{New mutation Operator}

\newacronym{rpd}{RPD}{Relative Percentage Differences}
\newacronym{nea}{NEA}{New Evolutionary Algorithm}
\newacronym{s-cluspt}{S-CluSPT}{}

\newacronym{hbrga}{HB-RGA}{Heuristic Based on Randomized Greedy Algorithm}
\newacronym{rga}{RGA}{Randomized Greedy Algorithm}
\newacronym{spta}{SPTA}{Shortest Path Tree Algorithm}
\newacronym{g-mfea}{G-MFEA}{a strategy for using multifactorial optimization}
\newacronym{k-mfea}{K-MFEA}{New Multifactorial Evolutionary Algorithm}

\newacronym{rim}{RIM}{Repairing Individual Method}
\newacronym{skdp}{CSC}{Construct the Solution of CluSPT}
\newacronym{skdp2}{CSC}{}
\newacronym{mpcx}{N-MPCX}{New Multi-Parent Crossover}

\newcommand{\scalefigure}{0.20}

\newtheorem{theorem}{Theorem}[section]

\newtheorem{lemma}[theorem]{Lemma}
\theoremstyle{definition}
\newtheorem{definition}{Definition}[section]

\begin{document}

\begin{frontmatter}
	
	\title{A bi-level encoding scheme for the clustered shortest-path tree problem in multifactorial optimization}
	
	\author[httb]{Huynh Thi Thanh Binh}
	\ead{binhht@soict.hust.edu.vn}
	
	\author[httb]{Ta Bao Thang}
	\ead{tabaothang97@gmail.com}
	
	\author[httb]{Nguyen Duc Thai}
	\ead{thaihawai@gmail.com}
	
	\author[pdt]{Pham Dinh Thanh\corref{cor1}\href{https://orcid.org/0000-0002-2550-9546 }{\includegraphics[scale=0.5]{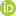}}}
	\ead{thanhpd05@gmail.com, thanhpd@utb.edu.vn}
	
	\cortext[cor1]{Corresponding author.}
	
	\address[httb]{School of Information and Communication Technology, Hanoi University of Science and Technology, Vietnam}
	\address[pdt]{Faculty of Natural Science and Technology, Taybac University, Vietnam}
	
	\begin{abstract}
		The \gls{cluspt} plays an important role in various types of optimization problems in real-life. Recently, some \glspl{mfea} have been introduced to deal with the \gls{cluspt}, but these researches still have some shortcomings, such as evolution operators only perform on complete graphs and huge resource consumption for finding the solution on large search spaces.  To overcome these limitations, this paper describes an \glsentrytext{mfea}-based approach to solve the \gls{cluspt}. The proposed algorithm utilizes Dijkstra's algorithm to construct the spanning trees in clusters while using evolutionary operators for building the spanning tree connecting clusters.  This approach takes advantage of both exact and approximate algorithms, so it enables the algorithm to function efficiently on complete and sparse graphs alike. Furthermore, evolutionary operators such as individual encoding and decoding methods are also designed with great consideration regarding performance and memory usage.  We have included proof of the repairing method's efficacy in ensuring all solutions are valid. We have conducted tests on various types of Euclidean instances to assess the effectiveness of the proposed algorithm and methods. Experiment results point out the effectiveness of the proposed algorithm existing heuristic algorithms in most of the test cases. The impact of the proposed \gls{mfea} was analyzed, and a possible influential factor that may be useful for further study was also pointed out.
	\end{abstract}
	
	
\end{frontmatter}

\glsresetall


\section{Introduction}
\label{Sec_Introduction}
In the years following massive globalization efforts, clustered problems have found great interest, not only within the confinement of research communities but even more so from members of governments, international enterprises, or streaming services. This need for an ordered structure in a world that grows increasingly closer in proximity comes about as a natural consequence, and so too arises the need to solve a problem whose ever-present trace can be found in this internationalized age. Evidently, there are more obvious examples of goods delivery within metropolitan areas and large-scale shipments between countries. There is also the problem of schedule planning for public transportation, optimization of the distribution network for streaming services, the arrangement of stores within shopping malls, and on the more abstract, conceptual side, structuring human resources of a project. One of the better known clustered tree problems is a variant of the shortest path tree problem, named \gls{cluspt}~\cite{demidio_hardness_2019}. Being NP-Hard, the preferred method to tackle this problem has mainly by means of approximation algorithms, as solving a large instance of the \gls{cluspt} using exact approaches is unfeasible and quite literally a waste of time.

A family of global optimization algorithms that has found considerable success in dealing with NP-Hard problems are \glspl{ea}~\cite{agoston_eiben_2015, thanh_new_2015}. These algorithms based their mechanisms on Darwin's theory of evolution and natural selection. Essentially, a multitude of solutions will first be randomized, encoded in a way that the solutions are susceptible to changes by evolution operators, namely mutation and crossover, which will be further elaborated later on in this paper. The quality of solutions, henceforth be called as individuals, should increase upon subsequent generation. While \glspl{ea} themselves have been subjects of research since the 1990s or, if one were to consider their more remote conception, the 1970s, this paper is based on a recent variation of the algorithm, the \gls{mfea}~\cite{huynh2020multifactorial, hanh2020evolutionary}. The \gls{mfea} operates on the same principle as the \gls{ea}, but with a few adjustments to let it solve multiple problems at once as opposed to the \gls{ea}. This fundamental change differs the \gls{mfea} greatly from its ancestor, using the fitness landscapes of multiple problems to help complement each problem's solution, a concept known as cultural transmission, whereas the \gls{ea} has to navigate this fitness landscape alone and thus more prone to stuck in local optimal. The \gls{mfea} has found great success in solving different types of problems due to this implicit genetic transfer between tasks in a multitasking environment~\cite{gong_evolutionary_2019, feng_explicit_2020, zhou_toward_2020}.

Accordingly, the \gls{mfea} has been used in many research dealing with the \gls{cluspt}~\cite{ThanhPD_TrungTB,ThanhPD_DungDA, thanh2020two, thanh2020efficient}, applying different evolutionary operators and mechanisms to the base algorithm. However, because the \gls{mfea} solves multiple problems concurrently, each one with their own restrictions and distinct search space, many of these researches could not confidently satisfy all these restrictions, even if all the problems the \gls{mfea} has to solve is the same type with different inputs. As a result, the final solution sometimes turns out to be an invalid one, incompatible with the original search space. This paper introduces an approach that is capable of remedying this fatal drawback. Furthermore, as this approach combines both the approximation aspect of the \gls{mfea} and the exact aspect of Dijkstra's algorithm~\cite{xu2007improved}, it also improves the ultimate solutions, when compared to past algorithms on the \gls{cluspt}. Finally, we test the algorithm's effectiveness when using multiple parents in the crossover process instead of the default two.

The major contributions of this work are as follows:

\begin{itemize}
	\item Develop a new solution representation for the \gls{cluspt}, thus reducing resource consumption. This enhancement becomes especially more noticeable for instances of large proportion, whose population grows exponentially with size to ensure randomness and diversity.
	
	\item Devise a novel two-level scheme to solve the nested structure of the \gls{cluspt}: tree containing sub-trees. This scheme divides the \gls{cluspt} into an approximation problem and an exact problem, resulting in better solutions compared to pure approximation approaches while reducing the computational time as opposed to exact approaches.
	
	\item Propose a method to repair invalid individuals, ensuring that proposed evolutionary operators always produce a valid solution. The maximum number of fixes on an individual is also provided and proven.
	
	\item Design a crossover operator that can work with multiple parents.
	
	\item Propose an effective memory-based method to calculate the cost of \gls{cluspt} solution to increase the algorithm's performance.
	
	\item Analyze the experimental results on a diverse range of test instances, with additional comparison to other algorithms, to demonstrate this proposed algorithm's efficacy.
\end{itemize}

The rest of this paper is organized as follows. Section~\ref{Notation_and_definitions} presents the notations and definitions used for formulating problem. Section~\ref{Sec_Related_Works} introduced related works. The proposed \gls{mfea} for the \gls{cluspt} is elaborated in Section~\ref{Sec_Proposed_Algorithm}. Section~\ref{Sec_Computational_results} explains the setup of our experiments and analyzes the computed results. The paper concludes in Section~\ref{Sec_Conclusion} with discussions on the future extension of this research.

\section{Problem definition and notations}
\label{Notation_and_definitions}

The \gls{cluspt}  is defined on a simple, connected and undirected graph $G(V,E,C,w, s)$ with a set of vertices $V$, a set of edges $E$, a set of clusters $C$, a weight matrix $w$, and a source vertex $s$, respectively. Set of vertices V is divided into $m$ clusters $C^j, j = 1...m$, $V = C^1 \cup  C^2 \cup ...\cup C^m$ and set of clusters $C =\{  C^1, C^2,...,C^m\}$. A solution to this problem is a spanning tree $T$ whose sum of routing cost from the source vertex $s$ to all vertices in $V$ is minimum and the subgraph induced by all vertices in each cluster is also a spanning tree.

Given a spanning tree $T$ of $G$, let $d_{T}(u, v)$ denotes the shortest path length between $u$ and $v$ on $T$. 
The \gls{cluspt} is defined as following:
\begin{center}
	\begin{tabular}{l p{12cm}}
		\hline 
		\multicolumn{2}{c}{\textbf{Clustered Shortest-Path Tree Problem}} \\ 
		\hline 
		\hline 
		\textbf{Input}:		&  - A weighted undirected graph $G = (V, E, w)$.\\
		&  - Vertex set $V$ is partitioned into $k$ clusters ${V_1, V_2, . . ., V_k}$.\\
		&  - A source vertex $s$.\\
		\hline
		\textbf{Output}:   	&  - A spanning tree $T$ of $G$\\
		&  - Sub-graph $T[V_i] (i = 1,\ldots, k)$ is a connected graph.\\
		\hline 
		\textbf{Objective}: & $\displaystyle \sum_{v \in  V} d_{T}(s,v) \rightarrow $ min\\
		& where $d_{T}(u, v)$ is the cost of shortest path from vertex $u$ to vertex $v$ on $T$.\\ 
		\hline 
	\end{tabular}
\end{center}
\medskip

\begin{figure*}[!htb]
	\renewcommand{\scalefigure}{0.35}
	\begin{minipage}{0.68\textwidth}
		\centering
		\begin{subfigure}{0.35\linewidth}
			\centering
			\includegraphics[scale=\scalefigure]{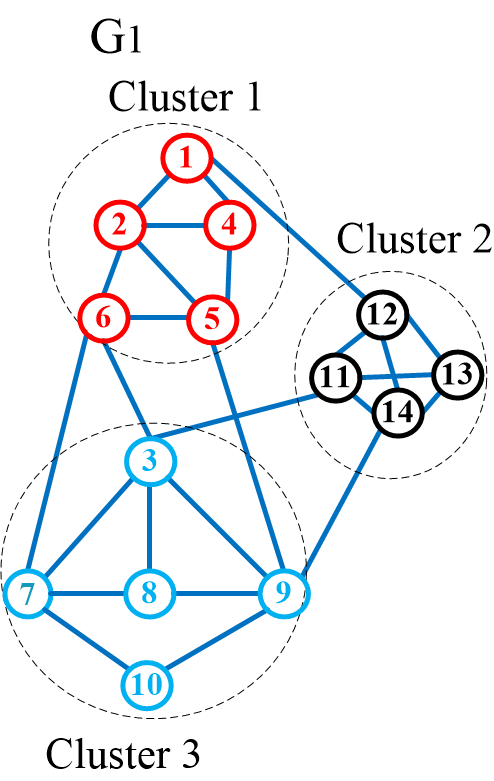}
			\caption{}
			\label{fig:inter-vertices_a}
		\end{subfigure}
		\begin{subfigure}{0.5\linewidth}
			\centering
			\includegraphics[scale=\scalefigure]{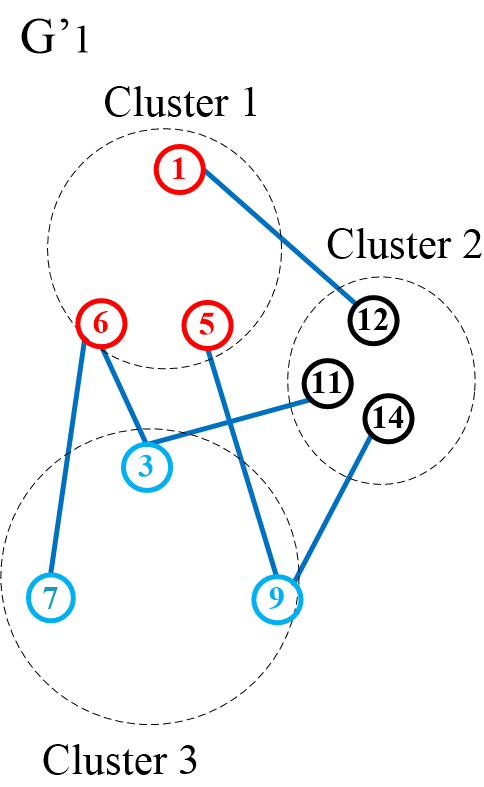}
			\caption{}
			\label{fig:inter-vertices_b}
		\end{subfigure}
		\caption{Sets of inter-vertices of the corresponding clusters in graph}\label{fig:inter-vertices}
	\end{minipage}\hfill
	\renewcommand{\scalefigure}{0.4}
	\begin{minipage}{0.3\textwidth}
		\centering
		\includegraphics[scale=\scalefigure]{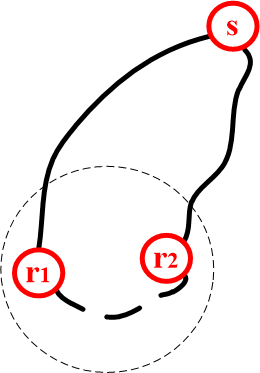}
		\caption{A spanning tree if exits a cluster having two local roots}\label{fig:my_proof}
	\end{minipage}
\end{figure*}

\begin{definition}
	For each cluster $C^j$ in $G$, $C^{j*}$ denotes the set of vertices connecting directly to vertices in other clusters. We call it the inter-vertices set of that cluster. An example is shown in Figure~\ref{fig:inter-vertices}. Figure~\ref{fig:inter-vertices_a} presents an initial input graph $G$ has 14 vertices and 3 clusters.  Figure~\ref{fig:inter-vertices_b} presents the set of inter-vertices of the corresponding cluster in $G$.
\end{definition}

\begin{definition}
	In the spanning tree $T$ obtained from the \gls{cluspt} problem, we denote a cluster's level by its proximity to the root cluster. The level of the cluster that contains the source vertex s is 0, the cluster directly connected to it is level 1, the cluster directly connected to a level 1 cluster is level 2,...etc. It is obvious that each non-root cluster has a single vertex serving as the entrance from a lower level cluster. We call this the local root of the cluster.
\end{definition}
\begin{lemma}
	Each cluster in the spanning tree T has only one local root.
\end{lemma}

\begin{proof}
	Assume that a cluster has more than one local root. Without loss of generality, we consider two local roots  $r_{1}$ and  $r_{2}$  of this cluster. There must exist a path  $p_{s,r_{1}}$ from the source vertex $s$ of $T$ to $r_{1}$ without going through $r_{2}$ and a path $p_{s,r_{2}}$ from  source vertex $s$ of $T$ to $r_{2}$  without going through $r_{1}$. In addition, the vertices $r_{1}$, $r_{2}$ belonging to the same cluster, so they will connect with each other, i.e there exists a path  $p_{r_{1},r_{2}}$ from $r_{1}$ to $r_{2}$. As a result, a cycle exists on $T$ illustrated in Figure~\ref{fig:my_proof}. This contradicts the requirement of $T$ being a spanning tree. The assumption is thus proven wrong.
	Therefore, each cluster in $T$ has only one local root meaning that every path from source vertex $s$ to vertices in a cluster in $T$ must go through its local root. Also, starting from $s$, a local root must be the first vertex to be visited within its own cluster. Because a local root is connected to a cluster of the lower level, the vertices selected as the local root of a cluster are in the cluster's inter-vertices set. It is for this reason that we have deemed only the inter-vertices necessary for being encoded. 
	
\end{proof}

\section{Related works}
\label{Sec_Related_Works}
Researches on cluster problems could be traced back to the 1970s~\cite{chisman_clustered_1975}. One of the earliest related studies is the Clustered Traveling Salesman Problem~\cite{bao_improved_2012, helsgaun_solving_2011}, whose objective is to determine the Hamiltonian path with a minimum length such that all vertices in each cluster are visited consecutively. As a cluster problem, the \gls{cluspt}~\cite{demidio_hardness_2019} has just recently been explored but has already attracted much interest from the research community. Because \gls{cluspt} is NP-Hard, exact algorithms for such problems require exponential execution time and are only suitable for small instances. Therefore the approximation approaches are more suitable to deal with the \gls{cluspt}.

In recent literature, \gls{mfea}~\cite{gupta_multifactorial_2016} has been emerging as an effective framework to solve a wide range of optimization problems. \gls{mfea} surpasses traditional \gls{ea}~\cite{pham2015effective} in its ability to simultaneously solve multiple problems, especially for systems with limited computational resources. Some preliminary works have been carried out, suggesting that \gls{mfea} is able to utilize the knowledge exchanging across relevant tasks to facilitate finding optimal solutions for multiple problems at the same time~\cite{bali_multifactorial_2019}.


Consider Parting Ways~\cite{wen_parting_2017}, the authors proposed a new mechanism to reallocate resources within the process of \gls{mfea}. This mechanism has stemmed from the observation that while exchanging genetic material across tasks is beneficial to the evolutionary algorithm's performance, but it can at times weaken exploitation and consume needless resources. More specifically, it is noted that the offspring generated from parents with different skill factors (called divergents) may help with the preservation of diversity within the population and allow information sharing to occur between tasks. However, as the population begins converging toward the optima of tasks, such contributions are irrelevant and, in cases, become obstacles for the convergence process. Therefore, a new measure ASRD, or Accumulated Survival Rate of divergents, was proposed to detect the point where information sharing is no longer desirable. In each generation, a ratio of divergents is created and survives to the next generation. The ASRD is the total of this ratio observed in a sliding window manner over several generations. Once ASRD is lower than a parameter value, the algorithm is set to never produce any divergent for the remaining duration.

According to Group-based~\cite{tang2018group}, the authors aimed to improve the \gls{mfea} by mean of a clustering-based approach.  It was discovered that in \gls{mfea}, the closer the global optima of two tasks are, or in general, the more similar the landscapes of two tasks are, the more likely it is for information sharing to have a positive impact. A new mechanism is built upon this discovery to maximize the advantage of information sharing in \gls{mfea}. More specifically, while the landscapes of tasks are mostly unknown, they can be roughly estimated by the distribution of the fittest individuals for each task. The new \gls{mfea} picks a representative for each task before comparing them using Manhattan distance, effectively clustering these representatives based on their likeness in genetic structure. Some representatives are further tested by exploring other clusters with regard to their skill factors. Should any cluster proved fruitful for the representatives, and more broadly for the task at which the representative is best at, improvement, then the representatives will be reassigned to such groups. Within one group, mating is only permitted if it could create better offspring. In addition, the authors also proposed a selection criterion to balance fitness value and diversity both in the mating and environmental selection (choosing individuals for the next generation) stage. This criterion is determined by the individual normalized fitness value and its crowding distance from its neighbors, together monitored by a balancing factor $\alpha$.

In Dynamic Resource Allocation~\cite{gong_evolutionary_2019}, the authors designed a way of dynamically reallocating resources for each task in \gls{mfea} and a method to control information sharing inside a particular task and across tasks. It is a fact that in solving multiple tasks, rarely comes the case where each task is of approximately equal hardness. Thus, it appears counter-intuitive to distribute computational resources equally among the tasks. The new mechanism split the population into subpopulation, each specializing in one task. A subpopulation consists of a specific number of fittest individuals for that task along with some random individuals, the number of whom monitored by a learning parameter that updates itself every iteration, whose participation serves to introduce cross-domain genetic information. All subpopulations then run through several generations to record the improvement of population $\omega_k$. This measurement is the ratio of change between a task's best fitness value of this generation and the one some generation before. The greater the ratio, that is, the more the best fitness value improves, the better the chance of a task being selected for resource allocation, denoted by Index of Improvement, or IoI. The IoI of a task is its improvement of the population over the total of all improvement of population. Thus, an IoI exists in the range of [0,1], and the selecting process is a Roulette one.

Yuan Yuan et al.~\cite{yuan2016evolutionary} investigated the use of \gls{mfea}, specifically on Permutation-based Combinatorial Optimization Problems. The authors noted that the use of random key encoding for chromosomes proved ineffective for this kind of problem and suggested another unified representation scheme and a new survivor selection method. Since the chromosome for each task is always a permutation sequence, the new representation has the unified chromosome be a sequence of the task with the highest dimension. For each task, it simply removes genes with values greater than the task's dimension while keeping the remaining genes in the same order. The introduced survivor selection method, dubbed as Level-based Selection, or LBS, sort the population into $K$ lists corresponding to $K$ optimization tasks. After the mating stage and all offspring have been put into the lists with regard to their skill factor, the list is sorted in ascending order (assuming, without loss of generality, that the tasks are minimization tasks), and an individual position in its list denotes its level. The population of the next generation is chosen from across all lists, and an individual can never be chosen if there is still another individual with a lower level in one of the lists.

Improvements upon the original \gls{mfea} have been met with considerable success. MFEA-II~\cite{bali_multifactorial_2019}, from the same authors, tackled the centralizing idea of the algorithm itself: the knowledge transfer between tasks. While individuals excelling in the same task can reproduce freely, on the intra-task crossover, \gls{mfea} allows only a fixed percentage of the population (this percentage denoted by random mating probability, or $rmp$). This paper thus described in detail a probability-based system to measure the effectiveness of this knowledge transfer. In essence, as each task still exists individually, that is, separated into their own solution spaces (and by mean of \gls{mfea} can they find a common representation), the intra-task learning process may prove itself an obstacle for reaching optimal solutions. This is where MFEA-II comes into play, an algorithm that can adapt and change accordingly the $rmp$ and, if need be, cut off knowledge transfer altogether. If previous \gls{mfea} algorithms have depended on a fixed $rmp$ (commonly 0.5), MFEA-II formulates $rmp$ as a $n \times n$ matrix, assuming the number of tasks is $n$. This matrix reflects the relationship between tasks, with $rmp_{a,b}$ denotes how well $task_a$ and $task_b$ influence each other. Naturally, the matrix is symmetrical, as the relation between $task_a$ and $task_b$ is the same as $task_b$ and $task_a$. As each task corresponds perfectly to itself, the main diagonal line is all 1. This $rmp$ matrix changes itself upon each iteration, each generation, eventually erasing bad relationships while promoting good ones. The MFEA-II has found great success on many different implementations, surpassing state-of-the-art strategies like \textit{separable natural evolutionary strategy} or \textit{exponential natural evolutionary strategy}.

In recent years, some approximation approaches were developed for solving \gls{cluspt}. In~\cite{ThanhPD_TrungTB}, the authors proposed \gls{mfea} (hereinafter E-MFEA) with new genetic operator algorithm. The major idea of this novel genetic operators is that first comes the construction of a spanning tree for the smallest sub-graph, afterward spanning trees for larger sub-graphs are created from the spanning tree of smaller sub-graphs. In~\cite{ThanhPD_DungDA}, the authors took advantage of the Cayley code to encode the solution of \gls{cluspt} and proposed genetic operators. The genetic operators introduced here are, conceptually, similar to the genetic operator for binary and permutation representations. However, it limits its application to complete graphs only. Therefore, the novel \gls{mfea}, too, is suitable exclusively for complete graphs. Binh at.el.~\cite{binh_new_2019} discussed a new algorithm based on the \gls{ea} and Dijkstra's Algorithm. In a divide and conquer fashion, the proposed algorithm decomposes the \gls{cluspt} problem into two sub-problems. The first sub-problem's solution is found by an \gls{ea}, while Dijkstra's Algorithm solves the second sub-problem. The goal of the first sub-problem is to determine a spanning tree that connects among the clusters, while that of the second sub-problem is to determine the best-spanning tree for each cluster. In~\cite{thanh2020efficient}, the authors described a method of applying \gls{mfea} based on deconstructing an original problem into two problems. In the proposed \gls{mfea}, the second task plays a role as a local search method for improving the solutions that are determined in the first task.

Although some \gls{mfea} algorithms were proposed for solving \gls{cluspt} problem in practice, they have revealed multiple drawbacks, i.e., only applicable on complete graphs, inefficient for finding the solution on large search spaces, each task only finds the solution of a different problem (differences in problem formulation or dimensionality) or the proposed evolution operators are only suitable for the \gls{cluspt} problem.  Therefore, to overcome these drawbacks, a new \gls{mfea} based approach is henceforth introduced to solve \gls{cluspt} problems.

\section{Proposed Algorithm}
\label{Sec_Proposed_Algorithm}
In this section, we introduce an approach based on \gls{mfea} with new encoding, decoding, and repairing schemes as well as new evolutionary operators to solve multiple \gls{cluspt} problems simultaneously. Each problem is  interpreted as a single task in the multitasking environment. The $i^{th}$ \gls{cluspt} task is performed on a input graph $G_{i}= ( V_{i},E_{i},w_{i} ,s_{i}),  i=1,\ldots,K$  where $V_{i},  E_{i}, w_{i}, s_{i}$  are set of vertices, set of edges, weight matrix and source vertex, respectively.  $V_{i}$ is divided into $m_{i}$ clusters. The $j^{th}$ cluster of  the $i^{th}$ task  is denoted by $C_{i}^{j} ,  j= 1,\ldots,m_{i}$ and  $C_{i} = \{ C_{i}^1, C_{i}^2, \ldots, C_{i}^{m_{i}} \}$. The local root of the $j^{th}$ cluster of  the $i^{th}$ task  is denoted by $r_{i}^{j}$.
The proposed algorithm's structure is presented in Algorithm~\ref{alg:mfea}, and the implementation steps of the algorithm are discussed in detail in the following subsections.

\begin{algorithm}
	\Begin
	{	
		$t \gets 0$\;
		{\small\tcc{Initialize initial population}	}	
		$P(0) \gets$ Randomly generate $N$ individuals from USS \Comment{Refer to Algorithm~\ref{alg:init_method}}\;		
		\ForEach{individual $p_i \in P(0)$}
		{
			Assign skill factor $\tau_i = i\%N +1$\;
			Construct the private representation $sr_i$ of individual $p_i$  in task $\tau_i$\Comment{Refer to Algorithm~\ref{alg:decoding1}}\;
			\uIf {$p_i$ doesn't have a valid representation}
			{
				Perform  the Repairing Individual Method on $sr_i$ \Comment{Refer to Algorithm~\ref{alg:repair}}\;
			}
			Construct a two-level solution of the \gls{cluspt} $p'_i$ based on $sr_i$ \Comment{Refer to Algorithm~\ref{alg:decoding2}}\; 
			Evaluate $p_i$ based on the \gls{cluspt} solution $p'_i$ for task $\tau_i$ only\; 
		}
		Update scalar fitness of each individual in $P(0)$\;
		
		\While{stopping conditions are not satisfied}
		{
			
			Offspring population $P_c(t) \gets \emptyset$\;
			\While{$|P_c(t)| < N$}
			{	
				Choose $k$ individuals $p_i (i = 1,\ldots,k)$ randomly from $P(t)$\;
				{\small\tcc{Perform crossover operator}}
				\uIf {(All selected individuals have same a skill factor) or ($rand < rmp$)}{
					$o_i (i=1,\ldots,k) \gets$ Perform the multi-parent crossover on $p_i (i =1,\ldots,k)$ \Comment{Refer to Algorithm~\ref{alg:crossover}}\;
					
					The skill factor of $o_i \gets$ Select randomly the skill factor of the parent\;
				}\Else{
					{\small\tcc{Perform mutation operator}}
					$o_i \gets$ Perform mutation on each individual $p_i(i = 1,\ldots,k)$ \Comment{Refer to Algorithm~\ref{alg:mutaion}}\;
					The skill factor of $o_i \gets$ the skill factor of the individual $p_i(i = 1,\ldots,k)$\;
				}
				
				{\small\tcc{Construct the \gls{cluspt} solution and evaluate the offspring}}
				Decode the private representation $sr_i$ of each individual $o_i$ \Comment{Refer to Algorithm~\ref{alg:decoding1}}\;
				
				\uIf{$sr_i$ is invalid}
				{
					Repair  $sr_i$ \Comment{Refer to Algorithm~\ref{alg:repair}}\;
				}
				Construct the \gls{cluspt} solution based on the $sr_i$\Comment{Refer to Algorithm~\ref{alg:decoding2}}\;
				
				Evaluate $o_i$ for task corresponding to assigned skill factor only\;
				$P_c(t) \leftarrow P_c(t) \cup \{o_i\}$, $i=1,\ldots,k$\;
			}
			
			$P_B(t) \gets$ the top 50\% best individuals from $P(t)$\;
			$R(t) \gets P_c(t) \cup P_B(t)$\;
			Update scalar fitness of each individual in $R(t)$\;
			$P(t+1) \gets$ Get $N$ fittest individuals from $R(t)$\; 			
			$t \leftarrow t+1$\;
		}
	}
	\caption{Proposed Algorithm to solve multiple the CluSPT problems}
	\label{alg:mfea}
\end{algorithm}

\subsection{Individual representation in the unified search space}
\label{subsec_uss}

The traditional \gls{ea} process has typically been focused on efficiently solving a single optimization problem at a time. Each solution for a problem has a private representation separately. In \gls{mfea}, the representations of solutions in particular problems are combined into the unified representation to reduce computational time and take advantage of the transfer of knowledge among tasks when solving them simultaneously. 

The \gls{uss} for $K$ tasks of the \gls{cluspt} is a graph $G_{u}(V, C, m)$ where:
\begin{itemize}
	\item The set of vertices $V$ of $G_u$ is partitioned into $m$ clusters where $m = \max( m_{1}, m_{2},\ldots, m_{K})$ and $m_{i}$, $i=1,\ldots,K$ is the number of clusters of the $i^{th}$ task.
	
	\item The $j^{th}$ cluster of $G_u$ is denoted by $C^{j}$, and $C^{j}$ = $C_{1}^{j*} \cup C_{2}^{j*} \cup \ldots \cup C_{K}^{j*}$ where $C_{i}^{j*}$ is the set of inter-vertices of the $j^{th}$ cluster in the $i^{th}$ task. 
	
	\item The set of clusters $C =\{ C^{1}, C^{2},\ldots, C^{m}\}$.
\end{itemize}

\renewcommand{\scalefigure}{0.31}
\begin{figure}[!htb]
	\begin{minipage}{0.75\textwidth}
		\centering
		\begin{subfigure}{0.35\linewidth}
			\centering
			\includegraphics[scale=\scalefigure]{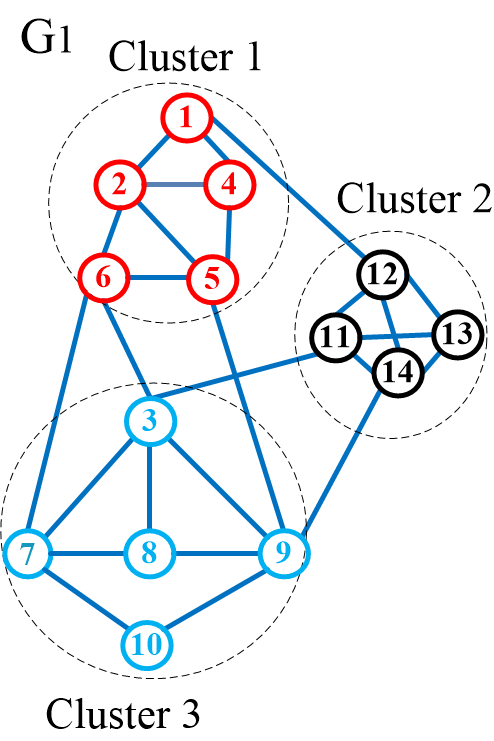}
			\caption{}
			\label{fig:unified_graph_a}
		\end{subfigure}
		\begin{subfigure}{0.35\linewidth}
			\centering
			\includegraphics[scale=\scalefigure]{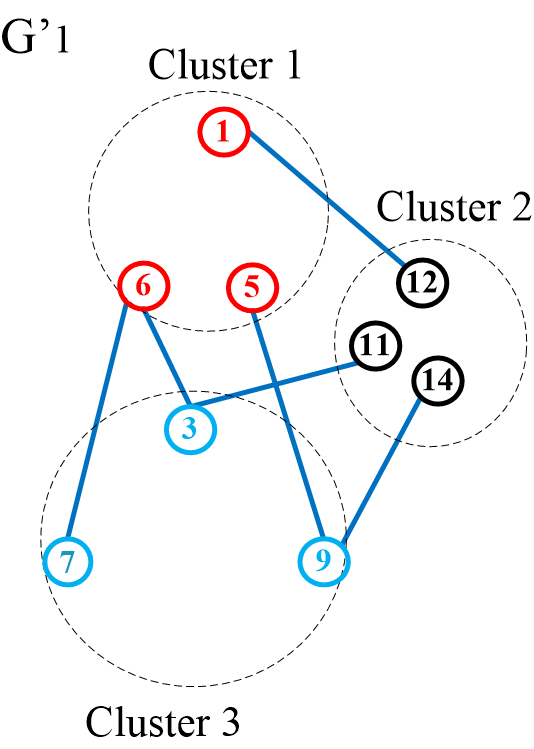}
			\caption{}
			\label{fig:unified_graph_b}
		\end{subfigure}
		\begin{subfigure}{0.35\linewidth}
			\centering
			\includegraphics[scale=\scalefigure]{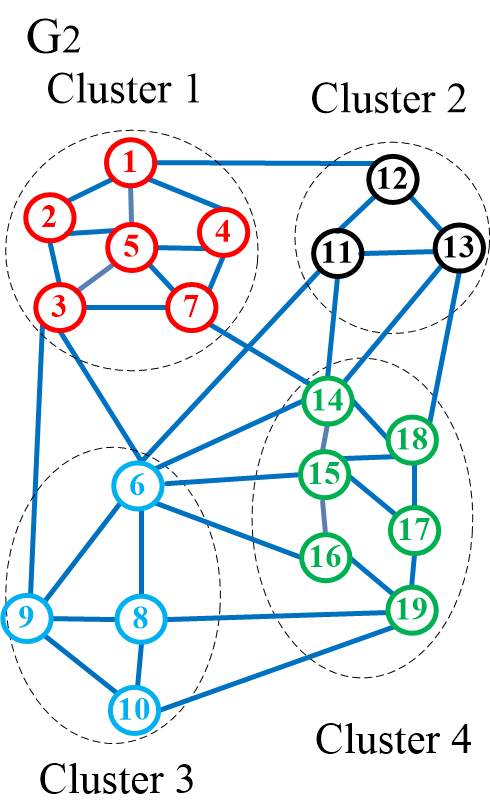}
			\caption{}
			\label{fig:unified_graph_c}
		\end{subfigure}
		\begin{subfigure}{0.35\linewidth}
			\centering
			\includegraphics[scale=\scalefigure]{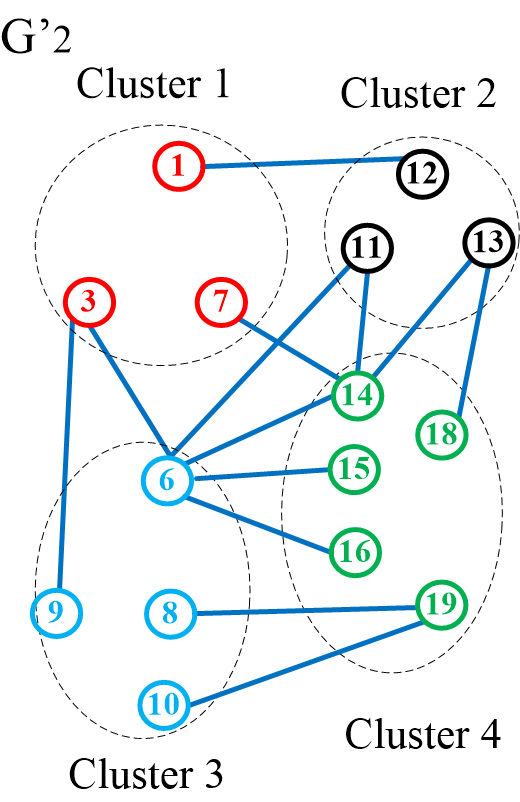}
			\caption{}
			\label{fig:unified_graph_d}
		\end{subfigure}
	\end{minipage}\hfill
	\begin{minipage}{0.25\textwidth}
		\begin{subfigure}{0.7\linewidth}
			\centering
			\includegraphics[scale=\scalefigure]{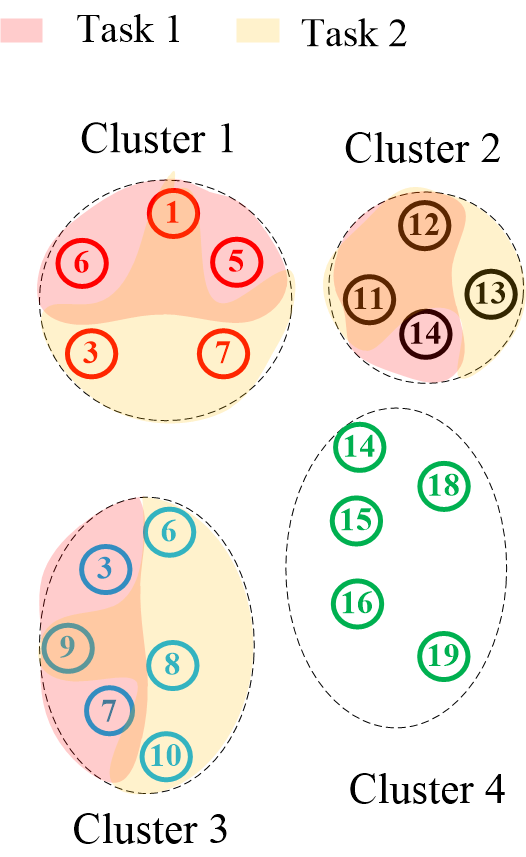}
			\caption{}
			\label{fig:unified_graph_e}	
		\end{subfigure}	
	\end{minipage}
	\caption{An example of a graph of the \glsentrytext{uss} for \gls{mfea} with two tasks}
	\label{fig:unified_graph}
\end{figure}

Figure~\ref{fig:unified_graph} illustrates steps to construct a graph $G_u(V, C, m)$ of the \gls{uss} from the graphs of two tasks $T_{1}$ and $T_{2}$. Figure~\ref{fig:unified_graph_a} and Figure~\ref{fig:unified_graph_c} describe the input graphs of the two tasks where graph $G_{1}$ contains 14 vertices while graph $G_{2}$ consists of 19 vertices. From the input graph of each task, we remove all vertices that do not directly connect to another cluster, leaving each cluster only with inter-vertices and their corresponding edges, resulting in the graphs $G_{1}^{'}$ and $G_{2}^{'}$, as shown in the Figure~\ref{fig:unified_graph_b} and Figure~\ref{fig:unified_graph_d}, respectively. The remaining vertices all fulfill the requirement of a local root. Next, we group vertices from different tasks by the index of their respective cluster. It should be noted that a vertex could potentially exist in different clusters of the graph $G_u$ because it represents a local root belonging to completely different tasks. Figure~\ref{fig:unified_graph_e} shows a graph of the \gls{uss} obtained, and the red area in each cluster denotes the vertices used for the first task while the orange area marked the vertices for the second task.

\setlength{\intextsep}{3pt}
\renewcommand{\scalefigure}{0.38}
\begin{figure}[htbp]
	\centering
	\includegraphics[scale=\scalefigure]{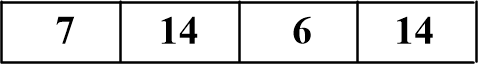}
	\caption{An example of an individual in the \glsentrytext{uss}}
	\label{fig:IndEncoding}
\end{figure}

An individual in the \gls{uss} is an array of vertices whose the $i^{th}$ element is a vertex belonging the $i^{th}$ cluster in $G_u$. Figure~\ref{fig:IndEncoding} illustrate an individual in the \gls{uss} which is created from the graph $G_u$ in Figure~\ref{fig:unified_graph_e}.

From this unified representation, we construct solutions for tasks through three main phases. Initially, a decoding phase is applied to find the private representation of the individual for each task. Afterward, the repairing phase will be conducted if needed in tasks so that a spanning tree can be built. Finally, a two-level construction strategy is conducted to build a \gls{cluspt} solution on the representation obtained after the above 2 phases. We will provide in-depth descriptions of the three phases in subsections~\ref{subsec_decode1},~\ref{subsec_repair} and~\ref{subsec_decode2}. 

\subsection{Individual Initialization Method}
\label{subsec_init}
With the advantage of the individual representation in the form of an array of integers, as shown in subsection~\ref{subsec_uss}, the initialization, crossover, and mutation methods can be done easily and effectively. Each element in the individual is selected randomly from the cluster of the graph $G_u(V, C,m)$ built in the previous section. The initialization method details are presented in Algorithm~\ref{alg:init_method} with time complexity of $O(m)$.

\begin{algorithm}[htbp]
	\KwIn{Graph $G_u(V,C,m)$} 
	\KwOut{An individual in the \glsentrytext{uss}}
	\BlankLine
	\Begin
	{	
		$I \gets \{ \}$ \;
		\For{k $\gets$  1 to m}
		{
			h $\gets$ Random a vertex in $C^{k}$\;
			I $\gets$ I $\cup$ \{h\}\;
		}
		\Return I\;
	}
	\caption{Initialization Individual Method}
	
	\label{alg:init_method}
\end{algorithm}

\subsection{Proposed Decoding Method}
\label{subsec_decode1}
This section describes a decoding method to construct an individual in each task from an individual in \gls{uss}. 

Each task's individual is an integer array whose dimension equals the number of clusters in that task. The $i^{th}$ element is used to determine the local root of the $i^{th}$ cluster unless said cluster contains the source vertex of the input graph, in which case that source vertex will be the local root of the cluster. In all other cases, let $l$ be the vertex corresponding to the $i^{th}$ element. If $l$ can be found within the cluster, we make it the local root. Otherwise, we locate the maximum index of vertex $l$ among the $i^{th}$ clusters of all tasks. This is a guarantee due to the way we initialize the chromosome. We then take the vertex whose index is the remainder of the division between the maximum index found earlier and the cluster's size. The method is described in Algorithm~\ref{alg:decoding1} with time complexity of $O(m^2)$, where $m$ is the number of clusters.

\begin{algorithm}[htbp]
	\setstretch{0.9}
	\KwIn{
		\begin{itemize}
			\item An  input graph of the $i^{th}$ task $G_{i}( V_{i}, E_{i}, C_{i},  C^{*}_{i}, s_{i},  m_{i})$,  $C_{i} =\{C^{1}_{i}, C^{2}_{i},..., C^{m_{i}}_{i}\}$ and  $C^{*}_{i} =\{C^{1*}_{i}, C^{2*}_{i}, ...,C^{m_{i}*}_{i}\}$.
			\item A graph $G_u(V,E,C,m)$.
			\item An individual  in the \gls{uss} $I =\{r_{1}, r_{2},..., r_{m}\}$ 
		\end{itemize}
	}
	\KwOut{An representation of individual in space search of the $i^{th}$ task}
	\BlankLine
	\Begin
	{
		$Genes \gets [ ]$\;
		\For{$j \gets 1$ to $m_i$}
		{
			
			$l \gets$ $r_j$   \Comment{ Value of the $j^{th}$ element in genes of individual $I$}\;
			\If{vertex l  doesn't appear in $C^{j*}_{i}$ of $G_i$}
			{
				$p \gets$  find maximum index of vertex $l$ in $C^{j*}_{h}$, $\forall h \in [1,\ldots,K]$ and ($h \ne i$)\;
				$size_j \gets$ the number of elements in  $C^{j*}_{i}$\;
				$p \gets$ $p \mod size_j$\;
				$l \gets$ value of the $p^{th}$ element in  $C^{j*}_{i}$\; 
			}
			$Genes$.append($l$)\; 
		}
		\Return $Genes$\;
	}
	\caption{Proposed Decoding Method}
	\label{alg:decoding1}
\end{algorithm}

\setlength{\intextsep}{3pt}
\renewcommand{\scalefigure}{0.35}
\begin{figure}[htbp]
	\centering
	\begin{subfigure}[b]{0.35\linewidth}
		\centering
		\includegraphics[scale=\scalefigure]{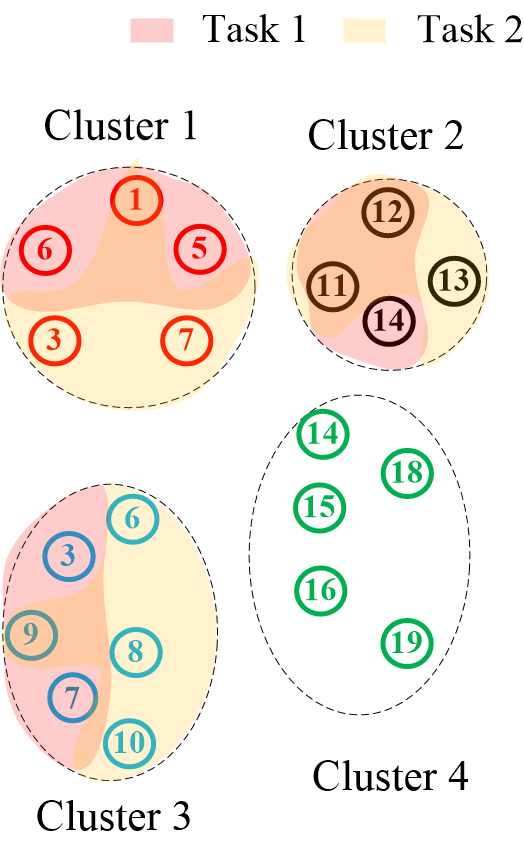}
		\caption{}
		\label{fig:decoding1_a}
	\end{subfigure}
	\begin{subfigure}[b]{0.5\linewidth}
		\centering
		\includegraphics[scale=\scalefigure]{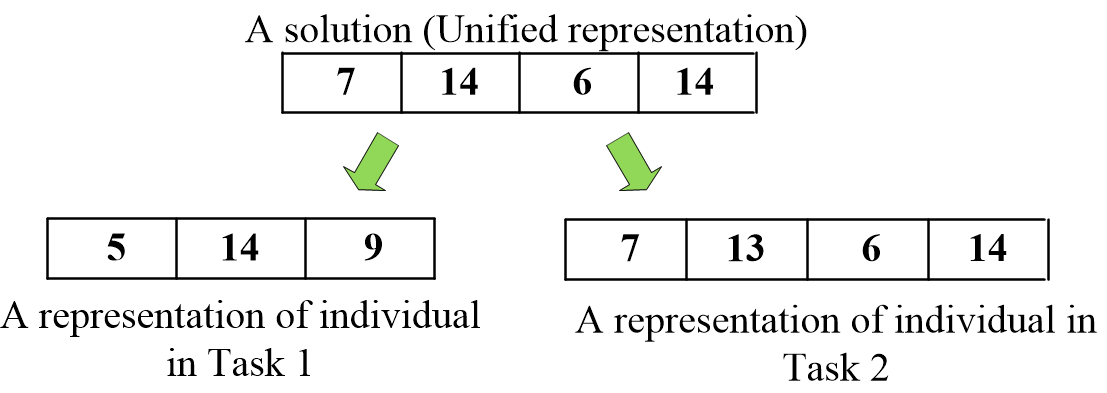}
		\caption{}
		\label{fig:decoding1_b}
	\end{subfigure}
	\caption{An example of steps of the decoding method to generate individuals for the tasks from an individual in \glsentrytext{uss}}
	\label{fig:decoding1}
\end{figure}

Figure~\ref{fig:decoding1} illustrates how the decoding method constructs the individuals in tasks $1$ and $2$ from the unified representation. Figure~\ref{fig:decoding1_a} provides a graph of the \gls{uss} $G_u = (V,C,m)$ with 4 clusters ($m = 4$). Sets of inter-vertices of clusters in Task 1 are $C^{1*}_{1} = \{1, 6, 5\}$, $C^{2*}_{1} =\{12, 11, 14\}$, $C^{3*}_{1} = \{9, 7, 3\}$ and in Task 2 are $C^{1*}_{2} = \{3, 1, 7\} $, $C^{2*}_{2} =\{11, 12, 13\}$, $C^{3*}_{2} = \{6, 8, 9, 10\}$, $C^{4*}_{2} = \{14, 15, 16, 18, 19\}$, respectively. The decoding process for Task 1 is as follow. Vertex 7 doesn't appear in $C^{1*}_{1}$ and the maximum index of vertex 7 in cluster 1 of other tasks (Task 2) is 2, so we choose a corresponding vertex to vertex 7 in the representation of Task 1. Vertex 5 is chosen since it has index $2$ ($= 2\mod3$) in cluster $C^{1*}_{1}$. Vertex 14 appears in $C^{2*}_{1}$, and thus it is added directly into the chromosome of Task 1. Vertex 6 doesn't exist within $C^{3*}_{1}$ and its maximum index in cluster 3 of the other task is 0, therefore vertex 9 having index $0$ ( $= 0\mod3$) will be chosen as the corresponding vertex in the representation. As a result, a representation of Task 1 is constructed. The process for Task 2 is similar. In Figure~\ref{fig:decoding1_b}, we demonstrate the two resulting individuals obtained after applying the decoding method.

\subsection{Repairing Individual Method}
\label{subsec_repair}
Although the individual representation as the proposed encoding method has many advantages in storing, calculating, and executing evolutionary operators, it contains a weakness that can be exploited in incomplete graphs. Because the inter-vertices that are elected as the root of clusters are randomly selected, they may not guarantee connectivity between clusters and violate the local root properties.  
An example  is depicted in Figure~\ref{fig:repair_a}.

In this example, Figure~\ref{fig:repair_a} presents the input graph of the task, Figure~\ref{fig:repair_b} shows a representation of an individual obtained after the first decoding stage, which doesn't guarantee local root properties of clusters. As shown in Figure~\ref{fig:repair_b}, vertex 12, 8, and 19 are assigned to be the local root of clusters 2, 3, and 4,  respectively. However, in the input graph, there is no path from 1 (source vertex) to 8 and 19, such that they must be the first visited vertices of their corresponding clusters, i.e., the local root property of the selected vertices cannot be guaranteed. 

Therefore, a \gls{rim} is introduced to fix this errors. Steps of \gls{rim} are as follows:
\begin{enumerate}
	\item[\textbf{Step 1}:] Add the cluster containing the source vertex to the closed set $V'$.
	\item[\textbf{Step 2}:] Among the remaining clusters, \gls{rim} adds into $V'$ all those whose local roots are connected to any vertex in $V'$.
	\item[\textbf{Step 3}:] If there are still clusters outside of V' after step 2 then do:
	\begin{enumerate}
		\item Randomize an edge $(u, v)$ with $v \in V'$ and $u \notin V'$.
		\item Determine the cluster containing $u$ and change the local root of that cluster into $u$.
		\item Add vertices of that cluster to set $V'$.
	\end{enumerate} 
	\item[\textbf{Step 4}:]  Repeat step 2 and step 3 until all clusters are added to $V'$.
\end{enumerate}

\setlength{\intextsep}{3pt}
\renewcommand{\scalefigure}{0.35}
\begin{figure*}[htbp]
	\centering
	\begin{subfigure}[b]{0.3\linewidth}
		\centering
		\includegraphics[scale=\scalefigure]{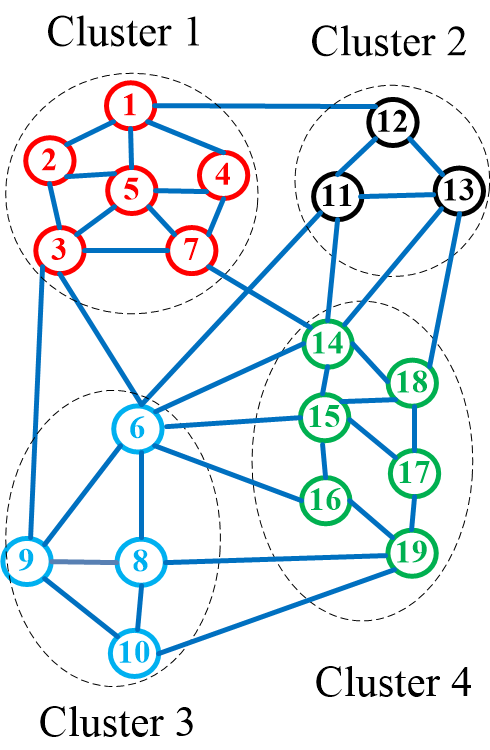}
		\caption{}
		\label{fig:repair_a}
	\end{subfigure}
	\begin{subfigure}[b]{0.3\linewidth}
		\centering
		\includegraphics[scale=\scalefigure]{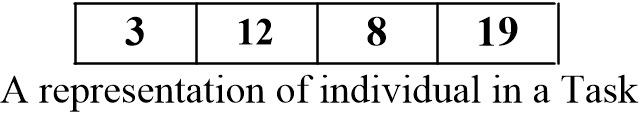}
		\caption{}
		\label{fig:repair_b}
	\end{subfigure}
	\begin{subfigure}[b]{0.3\linewidth}
		\centering
		\includegraphics[scale=\scalefigure]{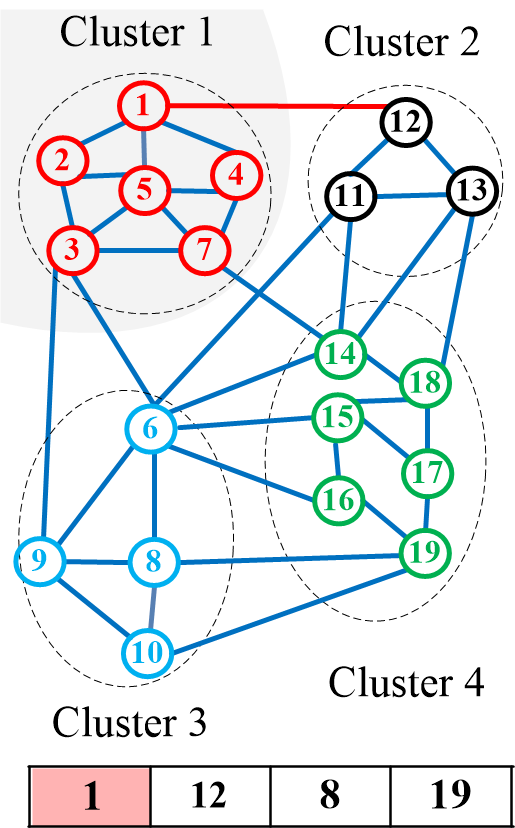}
		\caption{}
		\label{fig:repair_c}
	\end{subfigure}
	\begin{subfigure}[b]{0.3\linewidth}
		\centering
		\includegraphics[scale=\scalefigure]{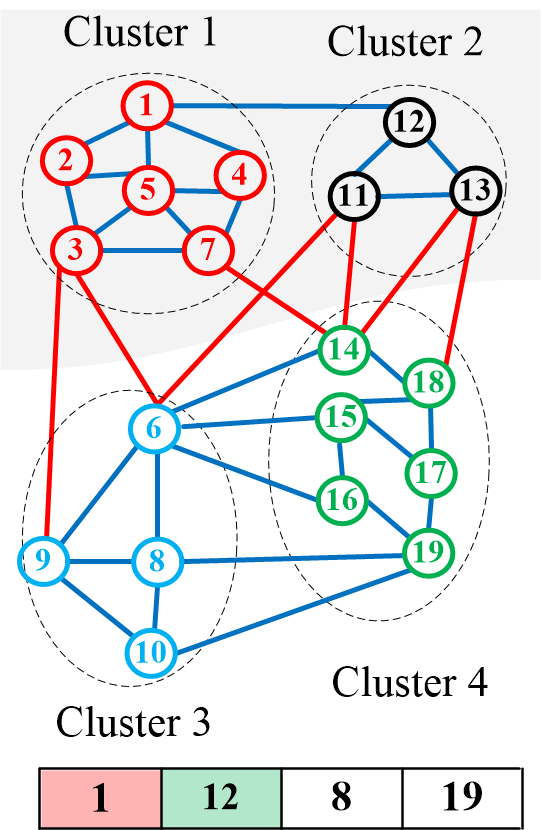}
		\caption{}
		\label{fig:repair_d}
	\end{subfigure}
	\begin{subfigure}[b]{0.3\linewidth}
		\centering
		\includegraphics[scale=\scalefigure]{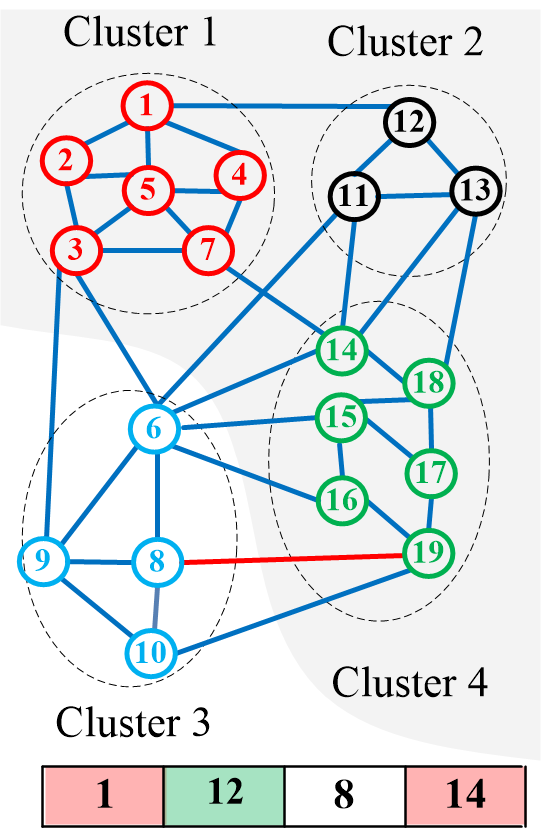}
		\caption{}
		\label{fig:repair_e}
	\end{subfigure}
	\begin{subfigure}[b]{0.3\linewidth}
		\centering
		\includegraphics[scale=\scalefigure]{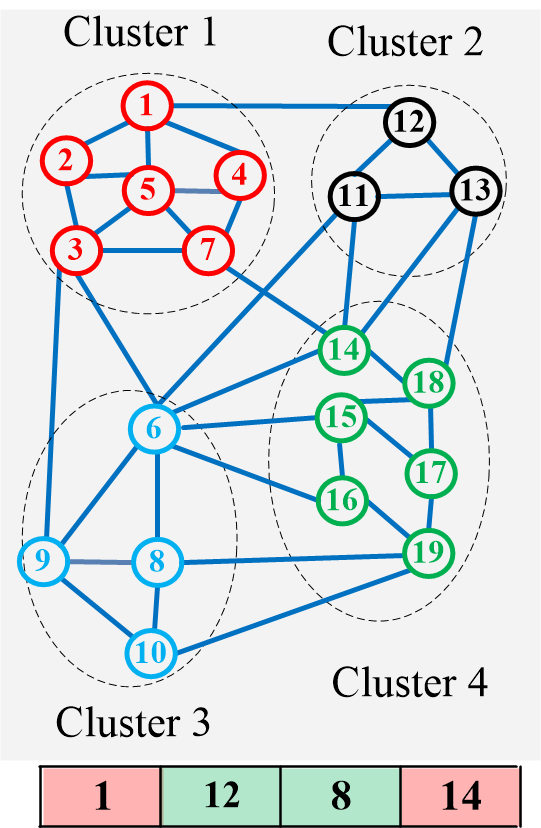}
		\caption{}
		\label{fig:repair_f}
	\end{subfigure}
	\caption{An example of steps to repair a individual representation}
	\label{fig:repairmethod}
\end{figure*}

\gls{rim}'s implementation is presented in Algorithm~\ref{alg:repair}. The algorithm's time complexity is $O(m*|C'|)$ where $m$, $C'$ are the number of clusters and vertices set of the graph in the \gls{uss}.

\begin{algorithm}[htbp]
	\setstretch{0.9}
	\KwIn{An input graph of the $i^{th}$ task $G_{i}( V_{i}, E_{i}, C_{i},  C^{*}_{i}, s_{i},  m_{i})$ where $C_{i} =\{C^{1}_{i}, C^{2}_{i},...,C^{m_{i}}_{i}\}$,  $C^{*}_{i} =\{C^{1*}_{i}, C^{2*}_{i}, ...,C^{m_{i}*}_{i}\}$.\\
		\qquad\quad An individual representation of the $i^{th}$ task $I = \{r_{1}, r_{2},...,r_{m_{i}}\}$.
	}
	\KwOut{A valid representation of $I$}
	\BlankLine
	\Begin
	{	
		$C^{m}_{i} \gets$ Determine the cluster containing the source vertex $s_{i}$ of $G_{i}$\;
		$I[m] \gets$ $s_{i}$\;
		$C' \gets \{C^{1*}_{i} \cup C^{2*}_{i}\cup \ldots \cup C^{m_{i}*}_{i}\}$\;
		$ V' \gets C^{m*}_{i}$\Comment{Add the cluster containing source vertex to  set $V'$}\;
		$visited[r_{j}] \gets false$ $\forall r_{j} \in I$ \;
		$visited[s_{i}] \gets true$\;
		$dem \gets m_{i} - 1$\;
		
		\While{$dem > 0$}
		{
			existsEdge $\gets$ false\;
			{\small\tcc{Add clusters whose local root connects directly to at least one vertex in V' to  V' }	}	
			\ForEach{$r \in I $}
			{
				\If{($visited[r] \ne true$) and ($\exists$ an edge $e = (r,k), k \in V'$)}
				{
					Determine cluster $C^{j*}_{i}$  containing vertex r \;
					$V' \gets V' \cup C^{j*}_{i}$\;
					$ dem \gets dem - 1$\;
					$visited[r] \gets$ true\;
					$existsEdge  \gets$ true\;
					break\;
				}
			}
			{\small\tcc{If no remaining cluster has an explicit connection to at least one vertex in V' through their local root, change the local root of a remaining random cluster }	}	
			\If{$existsEdge  = false$}
			{
				Select randomly an edge $e=(h,k)$,  $h \in V'$ and $k$ $\in C'\setminus V'$\;
				Determine cluster $C^{j*}_{i}$  containing vertex $k$\;
				$V' \gets C^{j*}_{i}$\;
				$dem \gets dem - 1$\;
				Replace the local root of cluster $C^{j}_{i}$ in $I$ by the vertex $k$\;
				$visited[k] \gets$ true\;
			}
			\Return $I$\;
		}
	}
	\caption{Repairing Individual Method}
	\label{alg:repair}
\end{algorithm}

Figures~\ref{fig:repair_c}--\ref{fig:repair_f} describe how the \gls{rim} is carried out. The chromosome has been decoded for this task, and the designated local roots are 3, 12, 8, and 19. As vertex 3 is in the same cluster as the source vertex 1, 1 is chosen as the first cluster's local root. In Figure~\ref{fig:repair_c}, we add cluster 1 into $V’$ and now only cluster 2 is directly connected to $V' $ through the edge (1,12) (highlighted in red). Thus, cluster 2 is added into $V' $, as shown in Figure~\ref{fig:repair_d}. Next, since both cluster 3 and 4 lack an explicit connection to $V' $ from their local roots, we instead randomize an edge satisfies this condition. It can be seen that the edges (3, 6), (3, 9), (6, 11), (18, 13), (14, 7), (14, 13) and (11, 14) are eligible for selection. Let the edge (11, 14) be chosen, we change the local root of cluster 4 into 14 and add cluster 4 into $V'$ (Figure~\ref{fig:repair_e}). Finally, we run through the clusters outside of $V' $ again, in this case only cluster 3, and find that it is connected to $V' $ using (8, 19). Cluster 3 is now added into $V' $, and the resulting local roots are now guaranteed to produce a spanning tree. 

Furthermore, to show \gls{rim} 's efficacy, we have provided and proven a lemma about the maximum number of positions to be fixed on the representation of each individual in Lemma~\ref{lamma:repair}.
\begin{lemma}
	\label{lamma:repair}
	The maximum number of positions that must be fixed on the genes is $\lfloor m/2 \rfloor$ with $m$ is the size of genes.
\end{lemma}
\begin{proof}
	Assume that we construct a graph G by considering a cluster as a vertex. An edge between two vertices $v_1$ and $v_2$ exists in G if cluster 1 connects to cluster 2 via its local root or vice versa. Because each element in representation is an inter-vertex that always has at least an edge connecting to another cluster's vertex (local root's property), each vertex in G is always connecting to another vertex. Therefore, in the worst case, $G$ is a forest with $\lfloor m/2 \rfloor$ connected components. To connect these connected components, it is necessary to modify one cluster's local root in each component. For that reason, the maximum number of positions in the representation that need to be corrected is $\lfloor m/2 \rfloor$.
\end{proof}	
\subsection{Construct a \glsentrytext{cluspt} solution for each task from representation of individual in its private search space}
\label{subsec_decode2}

To evaluate an individual in a task, we need to construct a \gls{cluspt} solution based on that individual. Therefore, this section describes a method (called \gls{skdp2}), which builds a \gls{cluspt} solution effectively from the corresponding individual in the task. The \gls{skdp2} method consists of two levels as follows:

\begin{itemize}
	\item[Level 1:] We will build the shortest path tree in each cluster of the graph from the local root of each cluster. The set of edges obtained is added to the set of edges of the solution in the task.
	
	\item[Level 2:]  We will build the connecting edges between the clusters by using a customized version of Dijkstra's algorithm. First, we add the entire set of vertices of the cluster containing the source vertex into a closed set $V$. Then we browse through the local root of the remaining clusters. Among the clusters connected to $V$ through their local roots, we search for the minimum route between a local root and the source vertex of the input graph. The cluster containing that root is then added to $V$, and the connecting edge is added to the edge set of the solution. Repeat until all clusters have been added to $V$. 
\end{itemize}

\setlength{\intextsep}{3pt}
\renewcommand{\scalefigure}{0.35}
\begin{figure}[htbp]
	\centering
	\begin{subfigure}[b]{0.3\linewidth}
		\centering
		\includegraphics[scale=\scalefigure]{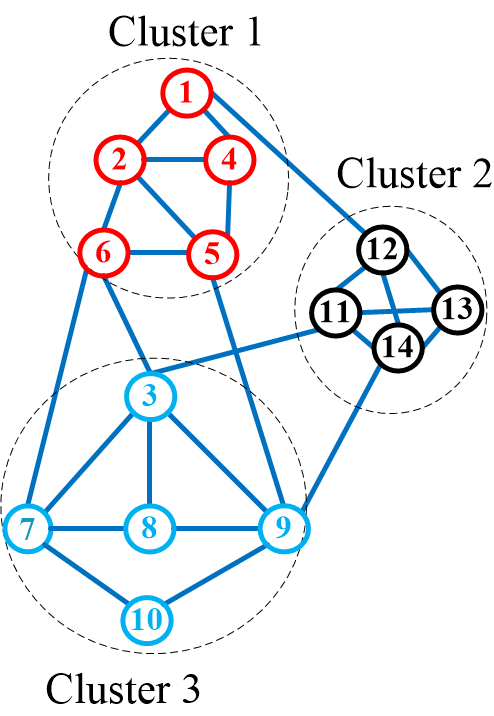}
		\caption{}
		\label{fig:decoding2_a}
	\end{subfigure}
	\begin{subfigure}[b]{0.3\linewidth}
		\centering
		\includegraphics[scale=\scalefigure]{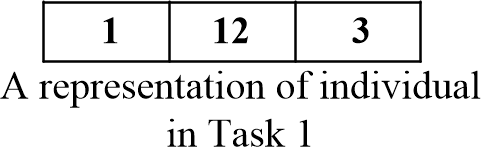}
		\caption{}
		\label{fig:decoding2_b}
	\end{subfigure}
	\begin{subfigure}[b]{0.3\linewidth}
		\centering
		\includegraphics[scale=\scalefigure]{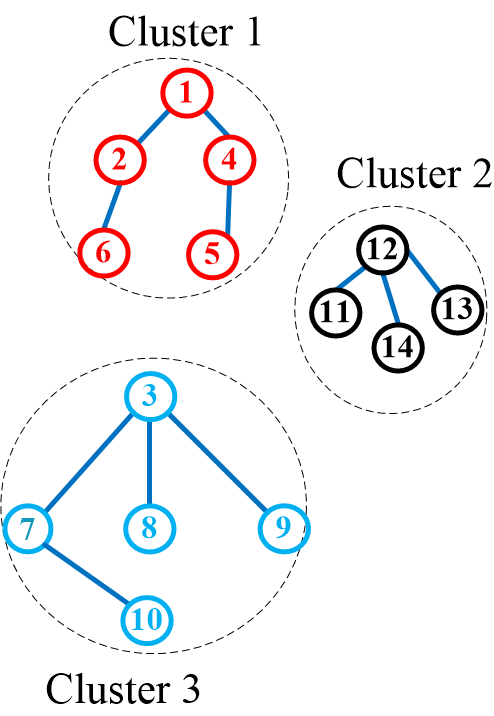}
		\caption{}
		\label{fig:decoding2_c}
	\end{subfigure}
	\begin{subfigure}[b]{0.3\linewidth}
		\centering
		\includegraphics[scale=\scalefigure]{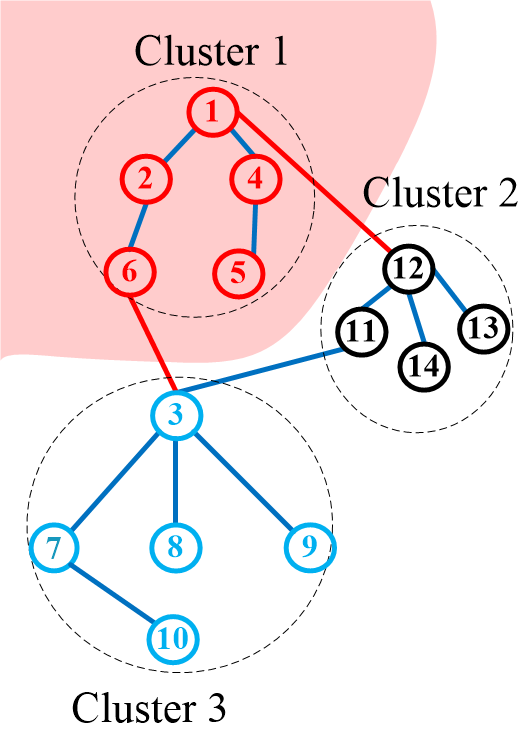}
		\caption{}
		\label{fig:decoding2_d}
	\end{subfigure}
	\begin{subfigure}[b]{0.3\linewidth}
		\centering
		\includegraphics[scale=\scalefigure]{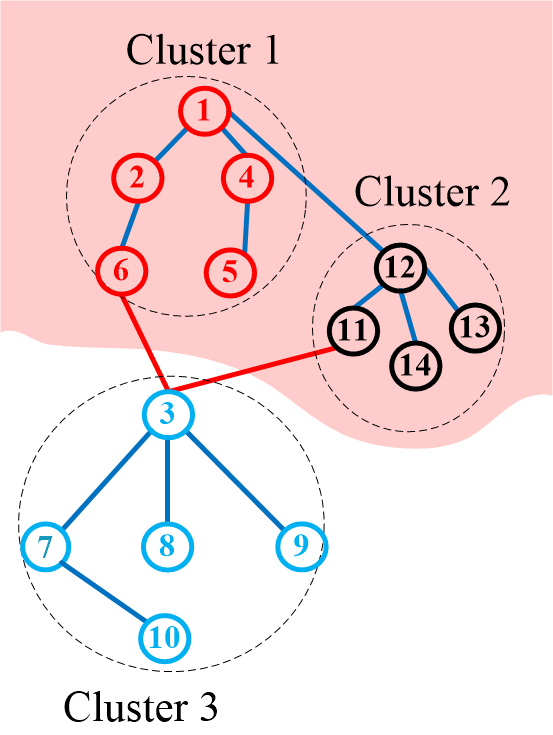}
		\caption{}
		\label{fig:decoding2_e}
	\end{subfigure}
	\begin{subfigure}[b]{0.3\linewidth}
		\centering
		\includegraphics[scale=\scalefigure]{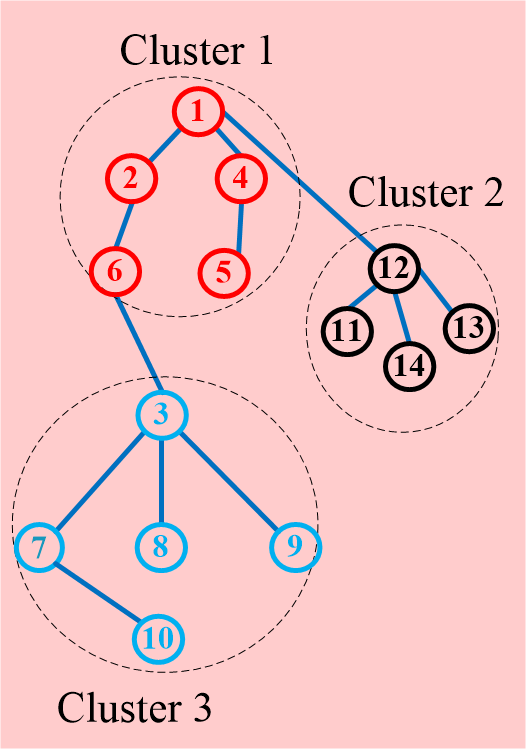}
		\caption{}
		\label{fig:decoding2_f}
	\end{subfigure}
	\caption{Steps to construct a solution from the private individual representation in the task}
	\label{fig:decoding2}
\end{figure}

The steps of the \gls{skdp2} is described in Algorithm~\ref{alg:decoding2}. The Dijkstra's algorithm in the paper is implemented using the Binary Heap structure and its complexity when running on the input graph $G(V, E)$ is $O(E+V*logV)$.  So for $m_{i}$ runs on $m_{i}$ clusters of graph  will cost $O (m_{i} * (E' + V'*log(V')))$ where $E'$, $V'$ are  edges and  vertices set of the largest cluster in the graph. In addition, the cost to build inter-cluster edges in $T$ is $O(m_{i}*V')$. Therefore, the complexity of the whole Algorithm~\ref{alg:decoding2} is $O(m_{i}*(E' + V' +V'log(V')))$.

\begin{algorithm}[htbp]
	\setstretch{0.75}
	\KwIn{
		An input graph of the $i^{th}$ task $G_{i}(V_i, E_i, C_i, s_i, m_i)$,  $C_i =\{C^1_i, C^2_i, ...,C^{m_{i}}_i\}$ and $C^*_i =\{C^{1*}_i, C^{2*}_i, ...,C^{m_{i}*}_i\}$.\\
		\qquad \quad An individual  $I = \{r_1, r_2,..,r_{m_{i}}\}$
	}
	
	\KwOut{A spanning tree $T(V, E)$ and sum of routing cost ($sum$)}
	\BlankLine
	\Begin
	{	
		$C^m_i \gets$ Determine the cluster containing source vertex $s_i$ of $G_i$\;
		$V \gets V_{i}$\;
		$E \gets \emptyset$\;
		$sum \gets$ 0 \Comment{the total routing cost from source vertex $s_i$ to all vertices in T}\;
		{\small\tcc{Construct the shortest path tree  in each cluster from its local root by Dijsktra Algorithm}	}
		\ForEach {cluster $C^j_i \in \mathcal C_i$}
		{
			$T_{j} \gets$ Determine the shortest path tree for $G_{i}[C^{j}_{i}]$ from local root $r_{j}$\;
			$d_{r_{j}k} \gets$  the cost of path from the local root $r_{j}$ to a vertex $k \in$ $C^{j}_{i}$ in $T_{j}$\;
			$sum \gets sum$ + $\sum d_{r_{j}k}, \forall k \in C^{j}_{i}$\;
			$E \gets E \cup E(T_{j})$\;
		}
		
		{\small\tcc{Construct edges connecting among clusters in T}	}
		$V^{'} \gets C^{m*}_i$\;
		$visited[r_{j}] \gets$ false\;
		$cost[r_{j}] \gets \infty, \forall r_j \in I$   \Comment{$cost[r_i]$ is the minimum cost to join the $j^{th}$ cluster into T through its local root $r_i$}\;
		$visited[s_{i}] \gets true$\;
		$dem \gets m_{i} - 1$\;
		{\small\tcc{Join cluster having the smallest additional cost}	}
		\While{$dem > 0$}
		{
			min $\gets \infty$\;
			$ v,u \gets$ $null$\;
			\ForEach{$ r \in I$}
			{
				\If{ $visited[r] \ne true$}
				{
					\ForEach{edge e=(r, k), k $\in$ V'}
					{
						$n_{cr} \gets$ the number of vertices in the cluster contain $r$\;
						$cost[r]$ $\gets$ min($cost[r]$,  ($d_{s_{i}k}$ + $c(k,r))*n_{cr}$)\;
						\If{ $cost[r] <  min$}
						{
							min $\gets$ $cost[r]$; u $\gets$ r\;
							v $\gets$ $arg\_min(cost[r])$\;
						}
					}
				}
			}
			sum $\gets$  sum + $cost[u]$ \Comment{Update the total routing cost}\;
			Determine cluster $C^{j}_{i}$  containing vertex $v$ and update routing cost from source vertex $s_{i}$ to vertices in $C^{j}_{i}$\;
			$V' \gets$  $C^{j*}_{i}$\;
			$E \gets$ $ E \cup (u,v)$\; 
			$dem \gets  dem-1$\;
			$visited[u] \gets$ true\;
		}
		\Return T(V, E), sum\;
	}
	\caption{Construct the Solution of CluSPT}
	\label{alg:decoding2}
\end{algorithm}

Figure \ref{fig:decoding2} presents a process of constructing the solution from the private representation of an individual in task $T_1$. Figure~\ref{fig:decoding2_a} shows the clusters of a task within the \gls{mfea} and in the Figure~\ref{fig:decoding2_b}, we see its individual representation. Vertex 1 is the source vertex of the input graph. Going through the chromosome, we get the three local roots: 1, 12, and 3. In the Figure~\ref{fig:decoding2_c} internal trees are created inside the clusters. We add the cluster containing source vertex  1 into $V$, as shown in Figure~\ref{fig:decoding2_d}. Next, we consider all the clusters that connect directly to $V$, in this case, both local roots 12 (through vertex 1) and 3 (through vertex 6). Comparing these two routes, we see that routing from 12 to 1 is closer than from 3 to 1, and thus cluster 2 is added to $V$ (figure \ref{fig:decoding2_e}). Finally, cluster 3 is connected to $V$ through both 6 and 11. Since the path $3\rightarrow 6 \rightarrow 1$ is shorter than the path $3 \rightarrow 11 \rightarrow 1$, we add cluster 3 into $V$ and the edge (3, 6) into the set of edges.

\subsection{Crossover Operator}
In traditional \gls{mfea}, the new children are produced from two parents who have the same skill factor or satisfy a random probability. In our proposed crossover operator (\gls{mpcx}), the parents have the same choice, but their children are produced from more parents in the hope that they could inherit much more characteristics of trees of their parent. This is similar to how farmers breed new crops that inherit the good traits of more than two-parent trees: good pest resistance of plant A,  high yield of plant B, good fruit quality of plant C, etc.

The \gls{mpcx} as shown in Algorithm~\ref{alg:crossover} and its time complexity is $O(N*m)$ with $N$, $m$ are number of parents and the number of genes.\\

Figure~\ref{fig:crossover} shows offsprings obtained after performing \gls{mpcx} 's steps. Figure~\ref{fig:crossover_a} presents 4 $(N=4)$ input parents and $N-1 = 4-1 = 3$ cut points selected randomly. The $i^{th}$ offspring preserves two segments of genes which are from the $(i-1)^{th}$ cut point to $i^{th}$ cut point and $(N-1)^{th} =3^{rd}$ cut point to the end of its corresponding parent. Other segments of the offsprings is determined by the gene segment of their next parents in order. If the segment selected is from $(i-1)^{th}$ cut point to $i^{th}$ cut point of the $i^{th}$ parent then change it to the next parent's gene segment in the same position. Figure~\ref{fig:crossover_b} shows 4 offsprings obtained.\\

\setlength{\intextsep}{3pt}
\renewcommand{\scalefigure}{0.4}
\begin{figure}[htbp]
	\centering
	\begin{subfigure}{0.48\linewidth}
		\centering
		\includegraphics[scale=\scalefigure]{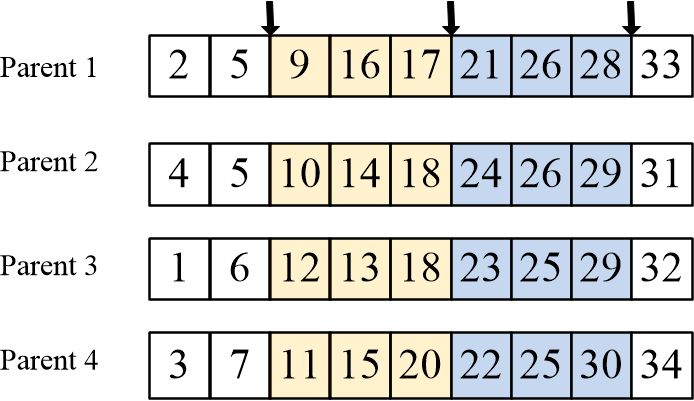}
		\caption{}
		\label{fig:crossover_a}
	\end{subfigure}
	\begin{subfigure}{0.48\linewidth}
		\centering
		\includegraphics[scale=\scalefigure]{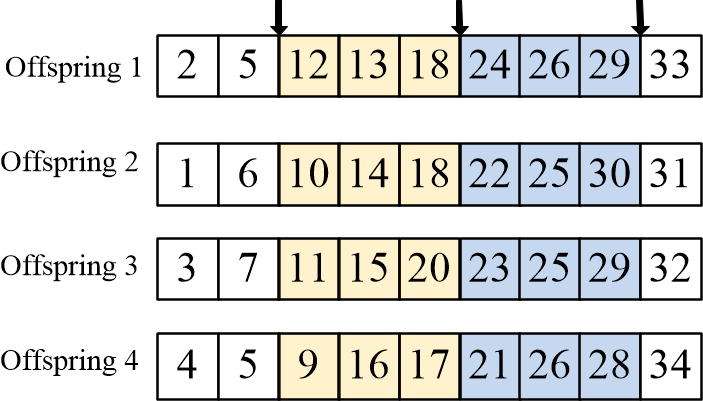}
		\caption{}
		\label{fig:crossover_b}
	\end{subfigure}
	\caption{An example for the proposed crossover operator}
	\label{fig:crossover}
\end{figure}

\begin{algorithm}[H]
	\setstretch{0.9}
	\KwIn{$N$ parents $P_{i} = \{p_{i1}, p_{i2},...,p_{im}\}$, $i=1,\ldots,N$, $2\leq N \leq m$.}
	\KwOut{$N$ offsprings $O_{i}$,  $i=1,\ldots,N$}
	\BlankLine
	\Begin
	{
		\uIf{$N=2$}{
			$cp_{i} \gets$ Select randomly two cut points where $1 \leq cp_{i} \leq m$ ($i=1, 2$), $cp_{1} < cp_{2}$\;
			$O_{i}$ $\gets$ $P_{i}$, $i = 1, 2$\;
			Swap genes from $cp_{1}$ to $cp_{2}$ of two offsprings $O_{1},O_{2}$ for each other\;
		}
		\Else{
			$cp_1,\ldots,cp_{N-1} \gets$ Select randomly $N-1$ cut points where ($1 \leq cp_i < cp_{i+1}\leq m-1$)\;
			$idx \gets$ 1 \;
			\For{i $\gets$  1 to N-1}{
				\For{j $\gets$  1 to N}{
					$h \gets j$\;
					\uIf{ $j\mod N+1 = i$}
					{
						$h \gets (j+1)\mod N+1$\;
					}\uIf{j $\ne$ i}{
						h $\gets$ $j\mod N+1$;
						
					}
					Copy the gene segment from index $idx$ to $cp_i$ in parent $P_{h}$ and append them into Offspring $O_{j}$\;
					
				}
				$idx \gets$ $cp_i +1$\;
			}
			\For{j $\gets$ 1 to N}{
				Copy the gene segment from index $idx$ to $m$ in parent $P_{j}$ and append them into Offspring $O_{j}$\;
			}
			
		}
		\Return {$O_{i}$, $i=1,\ldots,N$}\;
	}
	\caption{Proposed crossover operator}
	\label{alg:crossover}
\end{algorithm}

\subsection{Mutation operator}

The mutation operator is performed by randomly selecting a cluster and replacing its local root with another inter-cluster vertex in that cluster. The mutation method is described in Algorithm~\ref{alg:mutaion} with time complexity is $O(1)$.

\begin{algorithm}[htbp]
	\caption{Proposed mutation operator}
	\label{alg:mutaion}
	\KwIn{The Graph $G_u(V,C,m)$ where $C= \{ C^{1}, C^{2},.., C^{m}\}$,\\
		\qquad \quad An individual $I =\{r_{1}, r_{2},..,r_{m}\}$}.
	\KwOut{A new individual $I^* = \{r^*_{1}, r^*_{2},..,r^*_{m}\}$ }
	\BlankLine
	\Begin
	{
		$I^* \gets I$\;
		k $\gets$ Choose a random position in $I$\;
		${r'}_{k} \gets$ Choose a random vertex in cluster $C^{k}$\;
		$ r^*_{k} \gets {r'}_{k}$\;
		\Return $I^*$\;
	}
\end{algorithm}

\subsection{New method to speed up the evaluation for the \glsentrytext{cluspt} solution }
\label{subSec:speed_up}

Given a spanning tree $T$ is a solution of the \gls{cluspt} problem, the cost of the solution as defined is the sum of the shortest path distances from the source vertex $s$ to all the vertices in the tree:
\begin{equation}
\begin{split}
cost(T) &= \sum_{u\in V} d_T(s,u) \\ 
&=\sum_{i=1}^m \sum_{v\in C^i} d_T(s,v)\\
&=\sum_{i=1}^m \sum_{v\in C^i}\left( d_T(s,r_i) + d_T(r_i, v) \right)\\
&=\sum_{i=1}^m \left( | C^i | * d_T(s,r_i) + \sum_{v\in C^i}d_T(r_i, v)\right)\\
\end{split}
\end{equation}

From the above formula after decomposition, it can be easily seen that to calculate the cost of the solution, we need to calculate two components:
\begin{itemize}
	\item The cost of path from source vertex $s$ to the local root vertices of clusters $r_i$: $d_T(s,r_i), i = 1,\ldots,m$.
	\item The sum of costs of the path from the local root $r_i$ of each cluster to all vertices in that cluster $\sum_{v\in C^i}d_T(r_i, v)$.
\end{itemize}
In the above equation, the second component $\sum_{v\in C^i}d_T(r_i, v)$ does not depend on the source vertex $s$, but only on the local root and the subgraph of that cluster. And the result of this component can be determined by Dijkstra's algorithm once we know the local root $r_i$  of that cluster. Therefore,  to increase the algorithm's speed,  we only calculate this value of each cluster for each local root only once on the first calculation and save it to memory. When later calculation with the same local root is called, the corresponding memory cell's value will be retrieved. This saves a lot of computational costs because the cost of each Dijkstra's algorithm implementation is quite expensive. And this is also the reason why the idea of the encoding method is only based on the local root of each cluster in this paper is generated.

\section{Computational results}
\label{Sec_Computational_results}
\subsection{Problem instances}
To the best of our knowledge, there have been no instances of the \gls{cluspt} made available to the public. For this reason, a set of test instances based on the MOM~\cite{mestria_grasp_2013} (called MOM-lib) is generated. Various algorithms are utilized in the MOM-lib~\cite{demidio_clustered_2016}, resulting in six distinct types of instances. They are categorized into two kinds regarding dimensionality: small instances, each of which has between 30 and 120 vertices, and large instances, each of which has over 262 vertices. The instances are suitable for evaluating cluster problems~\cite{mestria_grasp_2013}.

However, to test the proposed algorithms' effectiveness in solving the \gls{cluspt}, it is necessary to add information about a source vertex to each instance. Therefore, a random vertex is selected as the source vertex for each instance.

For evaluation of the proposed algorithms, instances with dimensionality from 30 to 500 are selected. All problem instances are available via~\cite{Pham_Dinh_Thanh_2018_Instances}.


\subsection{Experimental criteria}
Criteria for assessing the quality of the output of the algorithms are presented in Table~\ref{tab:Criteria}.

\setlength{\intextsep}{3pt}
\noindent
\begin{table*}[!htb]
	\caption{Criterias for assessing the quality of the output of the algorithm}
	\centering
	\begin{tabular}{p{5.5cm} p{8.5cm}}
		\hline
		Average (Avg)    & The average function value over all  \\
		\addlinespace 
		Best-found (BF)       & Best function value achieved over all runs     \\
		\addlinespace  	
		\gls{rpd}                            & The difference between the average costs of two algorithms   \\
		\hline
	\end{tabular}
	\label{tab:Criteria}
\end{table*}

\setlength{\intextsep}{1pt}

In order to compare the quality of the \gls{cluspt} solutions received from distinct algorithms, \gls{rpd}~\cite{pop_two-level_2018} is used to calculate the difference between the average costs of two algorithms. The \gls{rpd} is calculated by the following formula:
\[
RPD=\dfrac{Solution - Best}{Best}*100\%
\]
, where $Best$ is the average cost of the solution obtained from the proposed algorithm, $Solution$ is the average cost of the solution obtained from existing algorithms, including \gls{hbrga} and \gls{g-mfea}.

The experimental results were recorded in tables given at the end of this paper. Each table was labeled with the numbers on the instances' types.

\subsection{Experimental setup}
To evaluate the performance of new \gls{mfea} for the \gls{cluspt}, two sets of experiments are implemented.
\begin{itemize}
	\item[$\bullet$] On the first set, two recent algorithms, \gls{g-mfea}~\cite{thanh2020efficient} and \gls{hbrga}~\cite{thanh_heuristic_2019} were implemented.
	\item[$\bullet$] On the second set, since the performance of the proposed \gls{mfea} were contributed by parameter: the relation between the number of vertices and the number of clusters,  R-experiment is conducted to evaluate the effect of these parameters. 
\end{itemize}

Each scenario was simulated 30 times on the computer (Intel Core i7 - 4790 - 3.60GHz, 16GB RAM), with a population size of 100 individuals evolving through 500 generations, which means the total numbers of task evaluations are 50000, the random mating probability is 0.5, and the mutation rate is 0.05. The source codes were installed by Java language.

\subsection{Experimental results}
To demonstrate the effectiveness of the proposed algorithm, we perform three analyses on the received results:
\begin{itemize}
	\item Statistical tests that have become a widespread technique in computational intelligence are first used to analyze the algorithms' performance.
	\item The details of comparison on each type of instance is discussed in the second analysis.
	\item In the third analysis, we point out factors that impact the performance of the proposed algorithm in comparison with existing algorithms.
\end{itemize}
\subsubsection{Non-parametric statistic for comparing the results of proposed algorithm and existing algorithm} \label{subsec:Statistic_singletask}

In order to examine the effect of the algorithms \gls{k-mfea}, \gls{hbrga}, and \gls{g-mfea} on the obtained results, we use Non-parametric for analyzing the received results. This study has two main steps:
\begin{itemize}
	\item The first step uses statistical methods such as Friedman, Aligned Friedman, Quade~\cite{derrac_practical_2011, carrasco_recent_2020} to test the differences among the results obtained by each algorithm.
	\item The second step is performed when the test in the first step rejects the hypothesis of equivalence of means, the detection of concrete differences among the algorithms can be made with the application of post-hoc statistical procedures~\cite{derrac_practical_2011, carrasco_recent_2020}, which are methods used for comparing a control algorithm with remaining algorithms.
\end{itemize}

Tables~\ref{tab:ResultsType1}-\ref{tab:ResultsType6} present the results obtained in the competition organized by types of instances and three algorithms. In these tables, the italic, red cells on a column in an algorithm denote instances where this algorithm outperforms two other algorithms.

\begin{table}
	\centering
	\caption{Results of the Friedman and Iman-Davenport test ($\alpha$=0.05)}\label{tab:Results_Friedman}
	\begin{tabular}{c c c c c c}
		\toprule
		Friedman Value & Value in $X^2$ & $p$-value & Iman-Davenport Value & Value in $F_F$ & $p$-value \\
		\cmidrule(l{3pt}r{3pt}){1-3} 
		\cmidrule(l{3pt}r{3pt}){4-6} 
		\textbf{175.353} & 5.991 & $9.869*10^{-11}$ & \textbf{235.745} & 3.028	 & $1.940*10^{-60}$\\
		\bottomrule
	\end{tabular}
\end{table}

The result of applying Friedman's and Iman-Davenport's tests is presented in Table~\ref{tab:Results_Friedman}. Given that the statistics of Friedman and Iman-Davenport are greater than their associated critical values, there are significant differences among the observed results with a probability error of $p \leq 0.05$.

\setlength{\intextsep}{6pt}
\begin{table}
	\centering
	\caption{Average rankings achiedved by the Friedman, Friednman Aligned, and Quade tests}\label{tab:Results_Rank}
\begin{tabular}{c c c c}
	\toprule
	Algorithms & Friedman & Friednman Aligned & Quade \\
	\midrule
	\gls{hbrga} & 2.669 & 273.831 & 2.600 \\
	
	\gls{g-mfea} & 2.209 & 230.939 & 2.323 \\
	
	 \gls{k-mfea} & 1.122 & 122.230 & 1.076 \\
	\bottomrule
\end{tabular}
\end{table}
\setlength{\intextsep}{6pt}

Table~\ref{tab:Results_Rank} summarizes the ranking obtained by the Friedman, Friedman Aligned, and Quade tests. Results in Table~\ref{tab:Results_Rank} strongly suggest the existence of significant differences among the algorithms considered.

The results in Table~\ref{tab:Results_Rank} show that the rank of the \gls{k-mfea} algorithm is the smallest, so we choose \gls{k-mfea} as the control algorithm. After that, we will apply more powerful procedures, such as Holm's and Holland's, to compare the control algorithm with the other algorithms. Table~\ref{tab:CompareControlAlg} shows all the possible hypotheses of comparison between the control algorithm and the remaining ones, ordered by their $p$-value and associated with their level of significance. 

\setlength{\intextsep}{6pt}
\begin{table}[!htp]
\centering
\caption{The z-values and p-values of the Friedman Aliged, Quade procedures (\glsentrytext{k-mfea} is the control algorithm)} \label{tab:CompareControlAlg}
\begin{tabular}{cc cccc cccc}
\toprule
& & \multicolumn{4}{c}{\textbf{Friedman}} &
\multicolumn{4}{c}{\textbf{Quade}}\\

\cmidrule(l{3pt}r{3pt}){1-2}
\cmidrule(l{3pt}r{3pt}){3-6}
\cmidrule(l{3pt}r{3pt}){7-10} 
$i$ & 
algorithms &
$z$ &
$p$ &
Holm &
Holland &
$z$ &
$p$ &
Holm &
Holland\\

\midrule
2 & \gls{hbrga} & 12.89 & $4.81*10^{-38}$ & 0.025 & 0.025 & 11.02  & $2.79*10^{-28}$ & 0.025 & 0.025\\
1 & \gls{g-mfea} & 9.05  & $1.35*10^{-19}$ & 0.05 & 0.05 & 9.02 & $1.85*10^{-19}$ & 0.05 & 0.05 \\
\bottomrule
\end{tabular}
\end{table}
\setlength{\intextsep}{6pt}

Table~\ref{tab:Adjust_p_FriedmanAliged} and Table~\ref{tab:Adjust_p_QUADE} show all the adjusted $p$ values for each comparison which involves the control algorithm. The $p$ value is indicated in each comparison, and we stress in bold the algorithms which are worse than the control algorithm, considering a level of significance $\alpha = 0.05$.

\setlength{\intextsep}{6pt}
\begin{table}[htbp]
\centering
\caption{Adjusted p-values for the Friedman test (\glsentrytext{k-mfea} is the control method)} \label{tab:Adjust_p_FriedmanAliged}
\begin{tabular}{c c ccccc}
\toprule
i & Algorithms & Unadjusted $p$ & $p_{Bonf}$ & $p_{Holm}$ & $p_{Hochberg}$ & $p_{Pli}$\\
\midrule
1 & \textbf{\glsentrytext{hbrga}} & $4.81*10^{-38}$  & $9.62*10^{-38}$ & $9.62*10^{-38}$ & $9.62*10^{-38}$ & $4.81*10^{-38}$\\
2 & \textbf{\glsentrytext{g-mfea}} & $1.34*10^{-19}$ & $2.70*10^{-19}$  & $1.34*10^{-19}$ & $1.34*10^{-19}$ & $1.34*10^{-19}$ \\
\bottomrule
\end{tabular}
\end{table}


\setlength{\intextsep}{6pt}

\setlength{\intextsep}{6pt}

\begin{table}[htbp]
\centering
\caption{Adjusted p-values for the QUADE test (\glsentrytext{k-mfea} is the control method)} \label{tab:Adjust_p_QUADE}
\begin{tabular}{ccccccc}
\toprule
i & Algorithms & Unadjusted $p$ & $p_{Bonf}$ & $p_{Holm}$ & $p_{Hochberg}$ & $p_{Pli}$\\
\midrule
1 & \textbf{\glsentrytext{hbrga}} & $2.79*10^{-28}$  & $5.59*10^{-28}$ & $5.59*10^{-28}$ & $5.59*10^{-28}$ & $1.84*10^{-28}$\\
2 & \textbf{\glsentrytext{g-mfea}}  & $1.85*10^{-19}$ & $3.70*10^{-19}$ & $1.85*10^{-19}$ & $1.85*10^{-19}$ & $1.85*10^{-19}$\\
\bottomrule
\end{tabular}
\end{table}
%

\setlength{\intextsep}{6pt}

\subsubsection{Detail of comparison among the algorithms \glsentrytext{k-mfea}, \glsentrytext{hbrga} and \glsentrytext{g-mfea} }
\paragraph{*) The aspect of solution quality}\mbox{}\\
\indent Tables~\ref{tab:SummaryResults} summarize the results obtained in comparison with two other algorithms. The results show that \gls{k-mfea} is significantly superior to \gls{hbrga}. \gls{k-mfea} exceeds \gls{hbrga} on most instances in all types, while the result received by \gls{hbrga} is equal to one received by \gls{k-mfea} on only an instance in Type 5 small. Between \gls{g-mfea} and \gls{k-mfea}, there are 29 test cases where the two algorithms' performance is equally matched. Combined with the results on Tables~\ref{tab:ResultsType1}-\ref{tab:ResultsType6}, we can see that a highlight in those test cases is the dimension of instances are often small, i.e., the dimension of instances in Type 1, Type 5, and Type 6 are less than 76, 65 and 76, respectively. \Gls{g-mfea} outperforms \gls{k-mfea} on two instances which are instance 5i500-304 in Type 5 Large and instance 9pr439-3x3 in Type 6 Large. However, \gls{k-mfea} surpasses \gls{g-mfea} on 108 out of 139 instances. Particularly, results obtained by the proposed algorithm are better than ones obtained by \gls{g-mfea} in all instances in Type 1 Large and Type 3.

\begin{table}[htbp]
  \centering
  \caption{Summary of comparison of results obtained by \glsentrytext{k-mfea} and the existing algorithms.}
    \begin{tabular}{cc rr rr r}
    \toprule
    & & \multicolumn{4}{c}{\textbf{K-MFEA}} &\\
    \cmidrule(l{3pt}r{3pt}){3-6}
          &       & 
          \multicolumn{1}{p{2cm}}{outperforms \glsentrytext{g-mfea}} & 
          \multicolumn{1}{p{2cm}}{is equal to \glsentrytext{g-mfea}} & 
          \multicolumn{1}{p{2cm}}{outperforms \glsentrytext{hbrga}} & 
          \multicolumn{1}{p{2cm}}{is equal to \glsentrytext{hbrga}} & 
          \multicolumn{1}{c}{\textbf{Total}} \\
   \midrule
    \multicolumn{1}{l}{\textbf{Type 1}} 
    	& Small & 18    & 8     & 26    & 0     & 26 \\
        & Large & 15    & 0     & 15    & 0     & 15 \\
	\addlinespace
    \multicolumn{1}{l}{\textbf{Type 3}} &       & 5     & 0     & 5     & 0     & 5 \\
    \addlinespace
    \multicolumn{1}{l}{\textbf{Type 4}} &       & 6     & 4     & 10    & 0     & 10 \\
    \addlinespace
    \multicolumn{1}{l}{\textbf{Type 5}} & Small & 10    & 10    & 19    & 1     & 20 \\
          & Large & 14    & 0     & 15    & 0     & 15 \\
    \addlinespace
    \multicolumn{1}{l}{\textbf{Type 6}} & Small & 28    & 7     & 35    & 0     & 35 \\
          & Large & 12    & 0     & 13    & 0     & 13 \\
    \bottomrule
    \end{tabular}%
  \label{tab:SummaryResults}%
\end{table}%

The average \gls{rpd} and the maximum \gls{rpd} of \gls{k-mfea} in comparison with \gls{g-mfea} and \gls{hbrga} on types are presented in Figure~\ref{fig:RPD}, in which the labels \gls{g-mfea} and \gls{hbrga} mean RPD(\gls{k-mfea}, \gls{g-mfea}) and RPD(\gls{k-mfea}, \gls{hbrga}) respectively.

In Figure~\ref{fig:RPD_Avg}, we can observe the average RPD(\gls{k-mfea}, \gls{g-mfea}) is larger than the average RPD(\gls{k-mfea}, \gls{hbrga}) on Type 6 small, Type 4 and both small and large instances on Type 1, and the biggest difference is on Type 4. The average RPD(\gls{k-mfea}, \gls{g-mfea}) in those types are of 4.4\%, 2.4\% and 4.1\% respectively. Conversely, the average RPD(\gls{k-mfea}, \gls{g-mfea}) is smaller than the average RPD(\gls{k-mfea}, \gls{hbrga}) on other types, in which the average RPD(\gls{k-mfea}, \gls{hbrga}) reaches its highest on Type 3 with the value of 2.2\% and its lowest on Type 5 small with the value of 1.1\%. Another remarkable point in Figure~\ref{fig:RPD_Avg} is that on Type 5 Small, the average RPD(\gls{k-mfea}, \gls{g-mfea}) is very low with the value of 0.04\%. It mean that on this type, the improvement of performance of \gls{k-mfea} is significantly small in comparison with \gls{g-mfea}. The reason behind this is that the dimension of instances on this type is small, leading to the difference between the two algorithms not being large.

Figure~\ref{fig:RPD_Max} indicates that the biggest gap between \gls{k-mfea} and \gls{g-mfea} is significantly larger than one between \gls{k-mfea} and \gls{hbrga}, and the maximum RPD(\gls{k-mfea}, \gls{g-mfea}) peaked at 18.6\% on Type 4. A notable point in this figure is that the smallest gaps also belong to RPD(\gls{k-mfea}, \gls{g-mfea}) with value of 0.32\% (Type 3), 0.3196\% (Type 5 small) and 0.334\% (Type 6 large). Smaller difference in this scene denotes that in comparison with \gls{k-mfea}, the results obtained by \gls{g-mfea} are closer than ones obtained by \gls{hbrga}. In other words, in comparison with \gls{g-mfea}, the solution quality produced by \gls{k-mfea} is not significantly improvement on these types.

\renewcommand{\scalefigure}{0.7}
\begin{figure*}[htbp]
	\centering
	\begin{subfigure}{.48\linewidth}
		\centering
		\includegraphics[scale=\scalefigure]{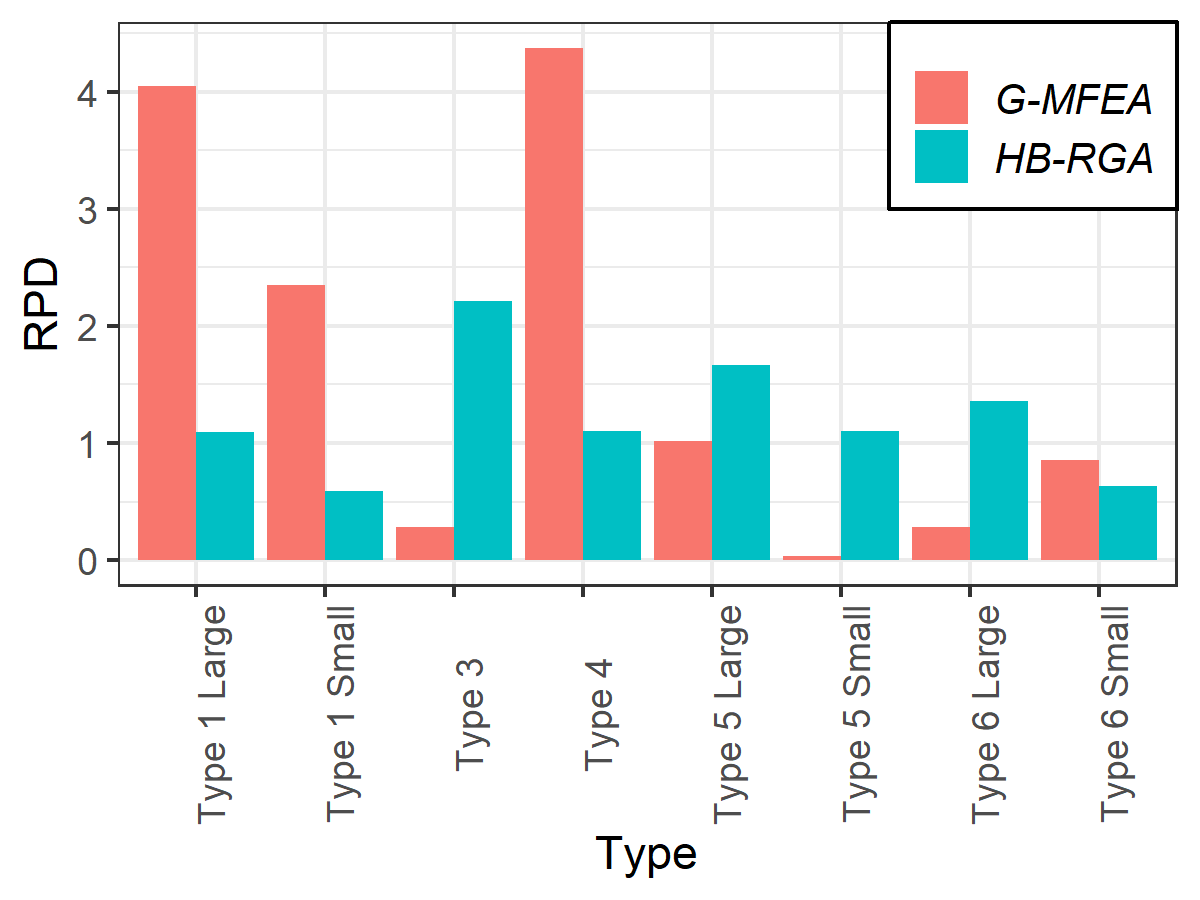}
		\caption{Average \glsentrytext{rpd}}
		\label{fig:RPD_Avg}
	\end{subfigure}
	\begin{subfigure}{.48\linewidth}
		\centering
		\includegraphics[scale=\scalefigure]{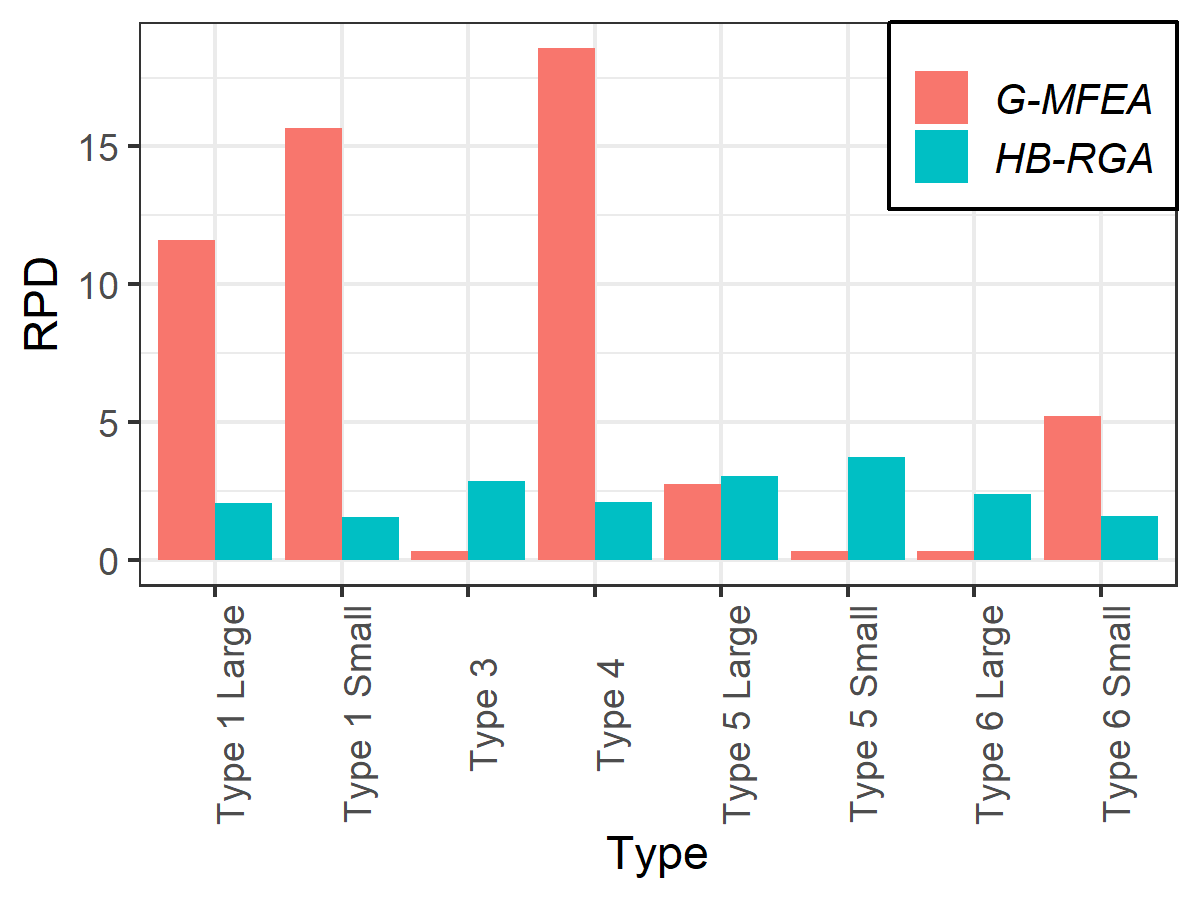}
		\caption{Maximum \glsentrytext{rpd}}
		\label{fig:RPD_Max}
	\end{subfigure}
	\caption{Average RPD and maximum RPD of \glsentrytext{k-mfea} in comparison with \glsentrytext{g-mfea} and \glsentrytext{hbrga} on types}
	\label{fig:RPD}
\end{figure*}

\paragraph{*) The aspect of running time}\mbox{}\\
\indent Figure~\ref{fig:RunningTime} depicts the average running time, in minutes, of the three algorithms. It can be seen quite clearly that \gls{g-mfea} is the inferior algorithm in this regard, compare to both \gls{hbrga} and \gls{k-mfea}. As such, the focus should be shifted to a comparison between \gls{hbrga} and \gls{k-mfea}, represented by Figure~\ref{fig:RunningTime2Alg}. For the small instances (Type 1 Small, Type 5 Small, Type 6 Small), \gls{k-mfea} achieves an execution time about 1.4-2 times faster than \gls{hbrga}. Excluding those types, however, \gls{k-mfea} has proven itself to be the superior method as \gls{hbrga} runs at least 6 times slower. This difference cannot be attributed alone to the nature of the \gls{k-mfea} and \gls{mfea} algorithms, which solve multiple problems at once, because \gls{g-mfea} too utilizes the \gls{mfea} model and loses out horribly on running time. A better explanation for this is the speed up method covered in this paper (section~\ref{subSec:speed_up}), which massively reduces computational stress.

\setlength{\intextsep}{6pt}
\renewcommand{\scalefigure}{0.6}
\begin{figure}[htbp]
	\centering
	\includegraphics[scale=\scalefigure]{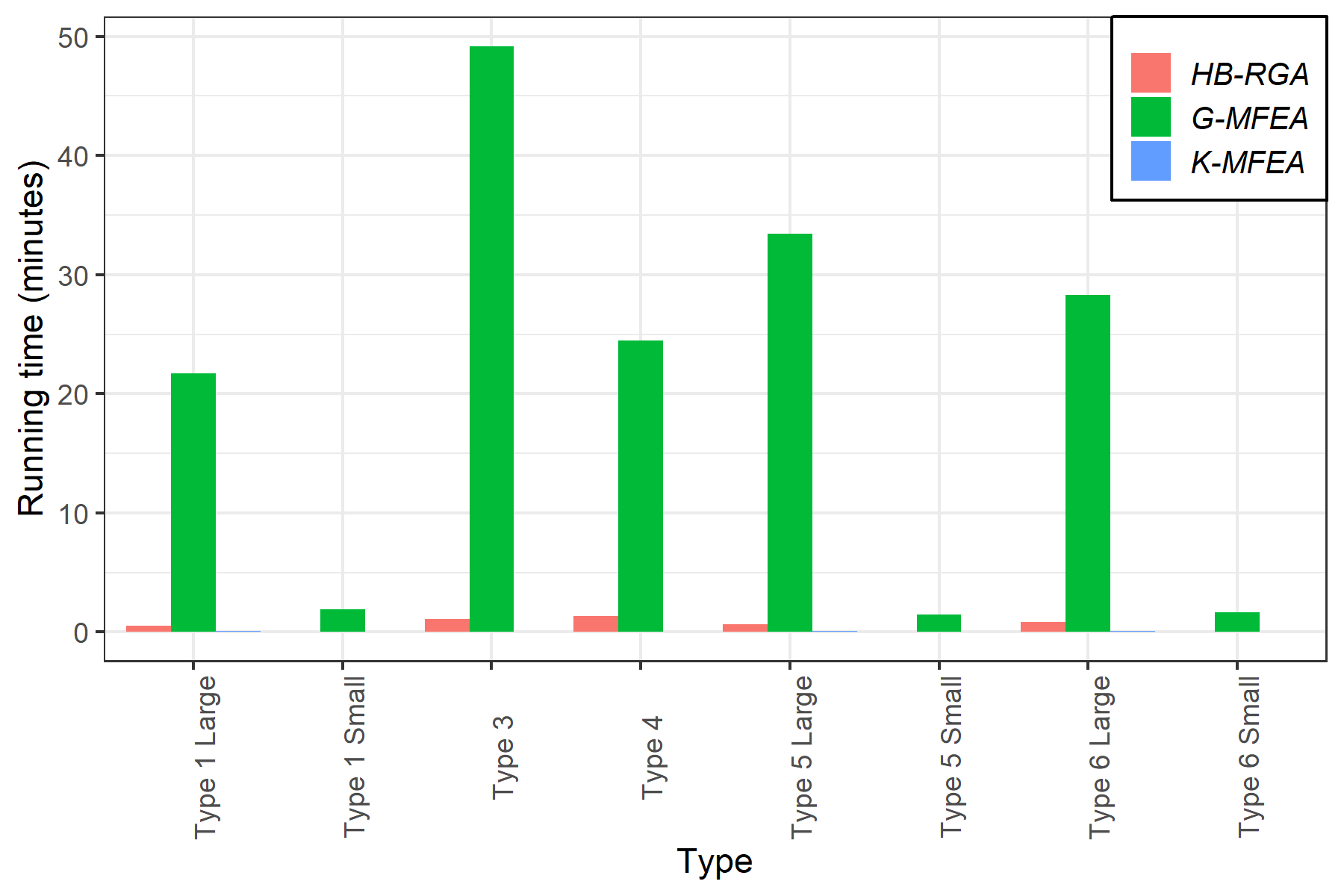}
	
	\caption{Running time of \glsentrytext{k-mfea}, \glsentrytext{g-mfea} and \glsentrytext{hbrga} on types}
	\label{fig:RunningTime}
\end{figure}
\setlength{\intextsep}{0pt}

\setlength{\intextsep}{6pt}
\renewcommand{\scalefigure}{0.6}
\begin{figure}[htbp]
	\centering
	\includegraphics[scale=\scalefigure]{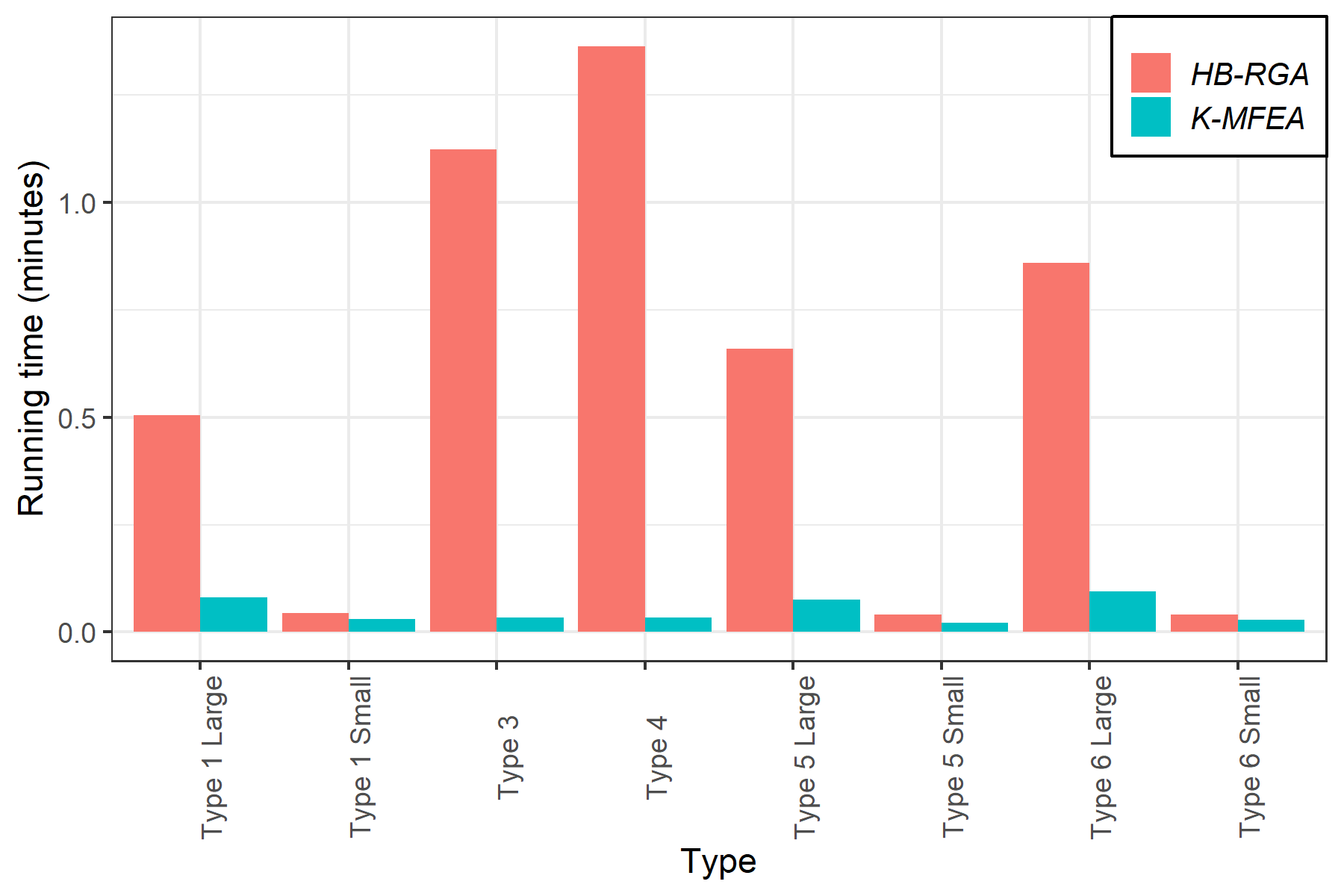}
	
	\caption{Running time of \glsentrytext{k-mfea} in comparison with \glsentrytext{hbrga} on types}
	\label{fig:RunningTime2Alg}
\end{figure}
\setlength{\intextsep}{0pt}

\subsubsection{Analysis of influential factors}

As mentioned in the previous subsection, the number of vertices may affect the performance of the \gls{k-mfea}. Therefore, In this subsection, the influence of the input graph's dimension (the number of vertices) on the performance of \gls{k-mfea} is analyzed. Because \gls{k-mfea} outperforms both \gls{g-mfea} and \gls{hbrga} on all instances of Type 3 and the number of instances in Type 4 is also very small, this study only considers the instances of Type 1, Type 5, and Type 6. 

In the previous subsection, the results have proven that \Gls{k-mfea} surpasses \gls{hbrga} in most test cases, i.e., on 138 out of 139 instances, so we only focus on analyzing the performance of \gls{k-mfea} in comparison with \gls{g-mfea}.

To determine the correlation between the number of vertices, the scatter plots of the relationship between the number of vertices and the comparison between two algorithms \gls{k-mfea} and \gls{g-mfea} for each type are graphed, and then this study tries to find the correlation coefficient in that relationship as shown in Figure~\ref{fig:Scatter_NumCluster}. In these Figures, circle symbols represent the performance of \gls{k-mfea} over \gls{g-mfea}, triangle means that the performance of \gls{g-mfea} outperforms \gls{k-mfea}, and square symbols illustrate the performance of the two algorithms when they are equal.

Figure~\ref{fig:Scatter_NumCluster} describes the relationship between the number of clusters and the performance comparison between \gls{k-mfea} and \gls{g-mfea}. A common feature of these figures is that the results from two algorithms are equal when the number of vertices is small, i.e., Type 1, Type 5, Type 6 is smaller than 76, 30, and 76, respectively. When the number of vertices is greater than 76 (for Type 1 and Type 6) and 30 (for Type 5), \gls{g-mfea} exceeds \gls{k-mfea} on only two test cases, which are on instance 5i500-304 in Type 5 and instance 9pr439-3x3 in Type 6. These lead to the conclusion that when the input graph's dimension increases, the performance of \gls{k-mfea} tends to increase.

\setlength{\intextsep}{0pt}
\renewcommand{\scalefigure}{0.6}
\begin{figure}[htbp]
	\centering
	\begin{subfigure}{.99\linewidth}
		\centering
		\includegraphics[scale=\scalefigure]{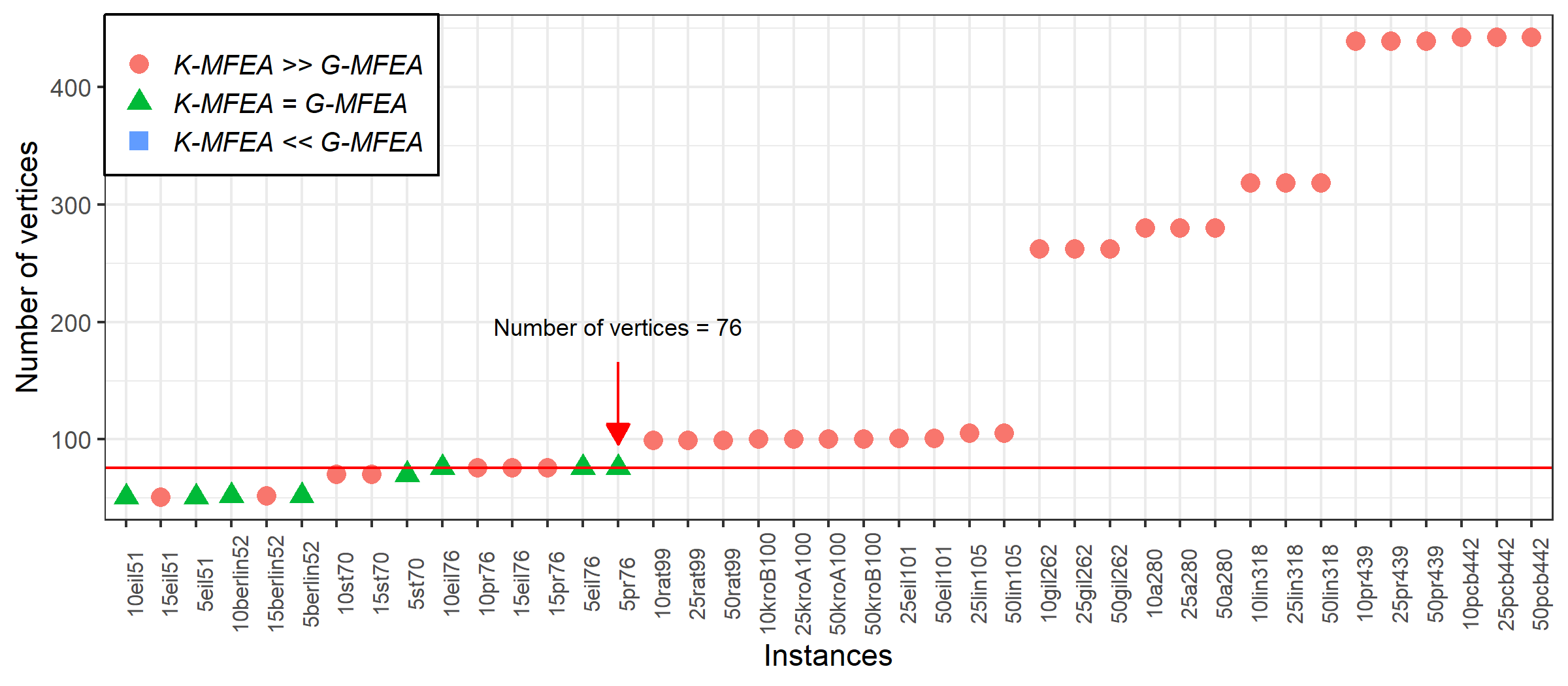}
		\caption{Type 1}
		\label{fig:Type1_NumCluster_N_MFEA_HBRGA}
	\end{subfigure}   
	\begin{subfigure}{.99\linewidth}
		\centering
		\includegraphics[scale=\scalefigure]{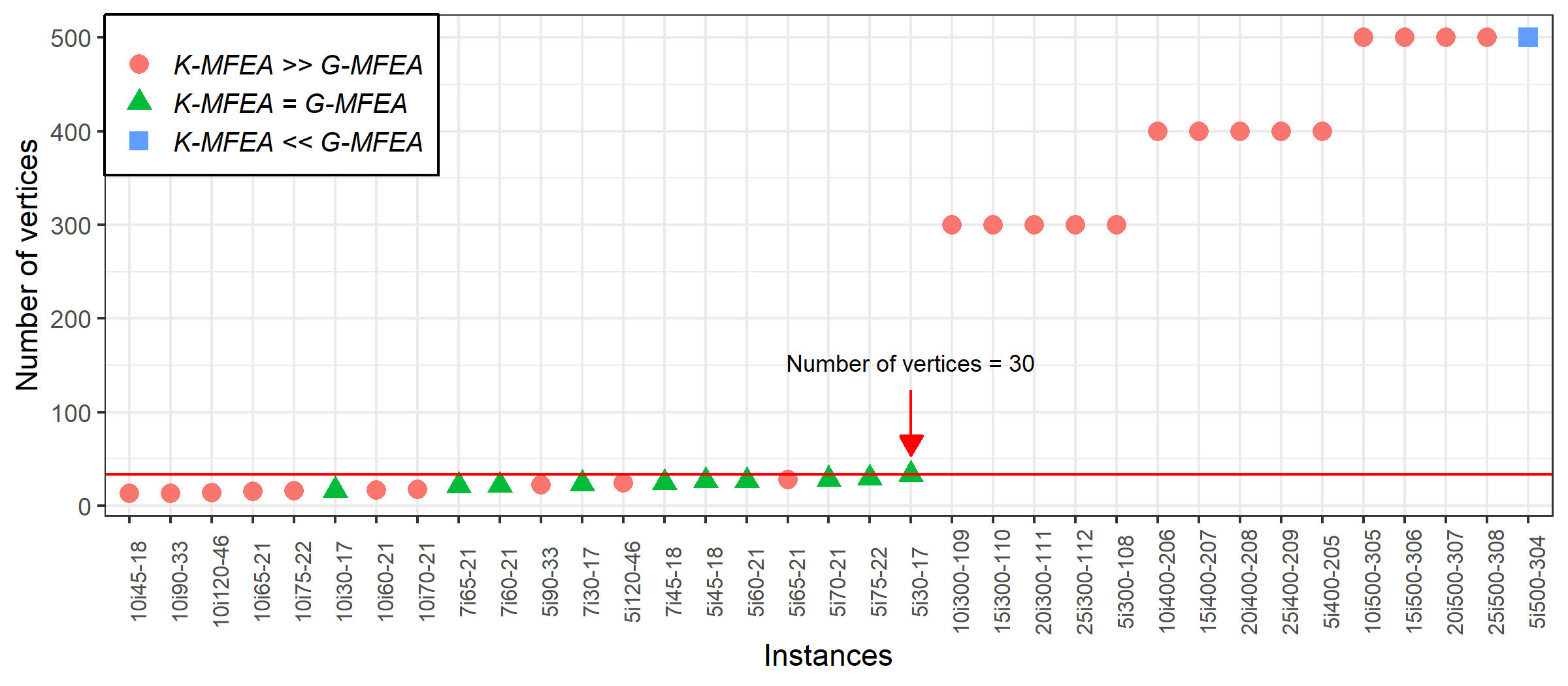}
		\caption{Type 5}
		\label{fig:Type5_NumCluster_N_MFEA_HBRGA}
	\end{subfigure}
	\begin{subfigure}{.99\linewidth}
		\centering
		\includegraphics[scale=\scalefigure]{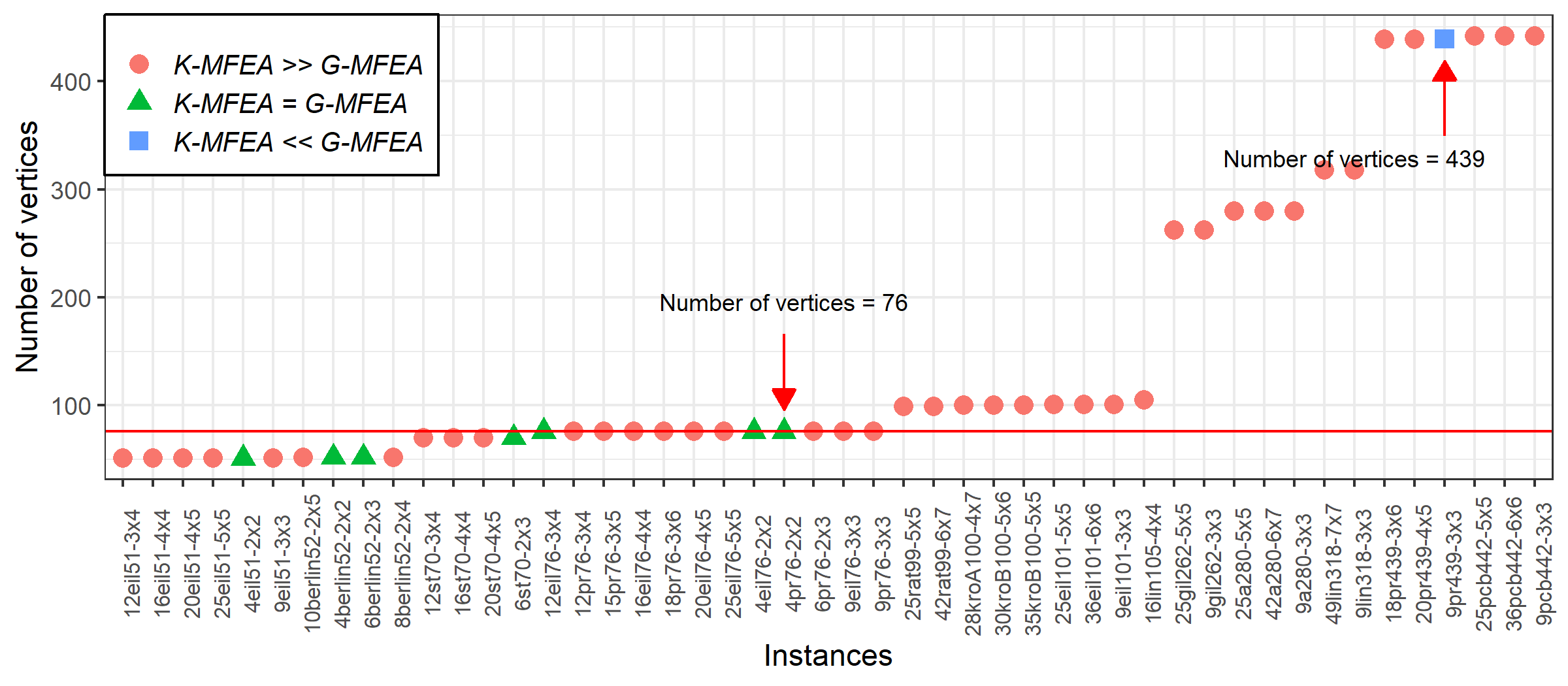}
		\caption{Type 6}
		\label{fig:Type6_NumCluster_N_MFEA_HBRGA}
	\end{subfigure}   
	\caption{The scatter of the number of vertices and the comparison between \gls{k-mfea} with \gls{g-mfea} in the Types 1, 5 and 6}
	\label{fig:Scatter_NumCluster}
\end{figure}

\subsubsection{Convergence trends}
\label{subsubsection:ConvergenceTrends}

In order to better understand the improved performance as a result of the proposed algorithm, we analyze the convergence performance of two algorithms \gls{g-mfea} and \gls{k-mfea} by using the functions in~\cite{gupta_multifactorial_2016} for computing the normalized objectives and averaged normalized objectives of the algorithms. In each type, a pair of instances are randomly selected for performing the analysis.

Figure~\ref{fig:ConvergenceTrends} illustrates the main convergence trends during the initial stages of \gls{k-mfea} and \gls{g-mfea} for instances 10pr439 and 25a280 in Type 1;  instances 5i300-108 and 5i400-205 in Type 5, and instances 9pcb442-3x3 and 9pr439-3x3 in Type 6. From Figure~\ref{fig:ConvergenceTrends} it can be noted that in general, the convergence trend of \gls{k-mfea} converges faster than one of \gls{k-mfea} and can categorize into two major trends:
\begin{itemize}
	\item The convergence trend of \gls{k-mfea} also surpassed that of \gls{g-mfea} on all generators as in Figure~\ref{fig:Type_1_convergence}.
	
	\item The convergence trend of \gls{k-mfea} surpasses that of \gls{g-mfea} on the last generators but on the first generators, the convergence of \gls{g-mfea} was often faster than that of \gls{k-mfea}. This trend is illustrate in Figure~\ref{fig:Type_5_convergence} and Figure~\ref{fig:Type_6_convergence}.
\end{itemize}

One other remark in Figure~\ref{fig:ConvergenceTrends} is the average normalized objective values difference between \gls{k-mfea} and \gls{g-mfea} at start time. The cause of this difference is that two algorithms encode the \gls{cluspt} solution by two different methods, so initial individuals of an algorithm differ from initial solutions of the rest algorithm. These observations led to the conclusion that the \gls{cluspt} solution produced by \gls{k-mfea} was often better than ones produced by \gls{g-mfea} in most generators. In other words, the evolution methods in \gls{k-mfea} may help to improve the quality of solution of the \gls{cluspt} in comparison with the methods in \gls{g-mfea}.


\setlength{\intextsep}{6pt}
\renewcommand{\scalefigure}{0.53}
\begin{figure}[htbp]
	\centering
	\begin{subfigure}{.32\linewidth}
		\centering
		\includegraphics[scale=\scalefigure]{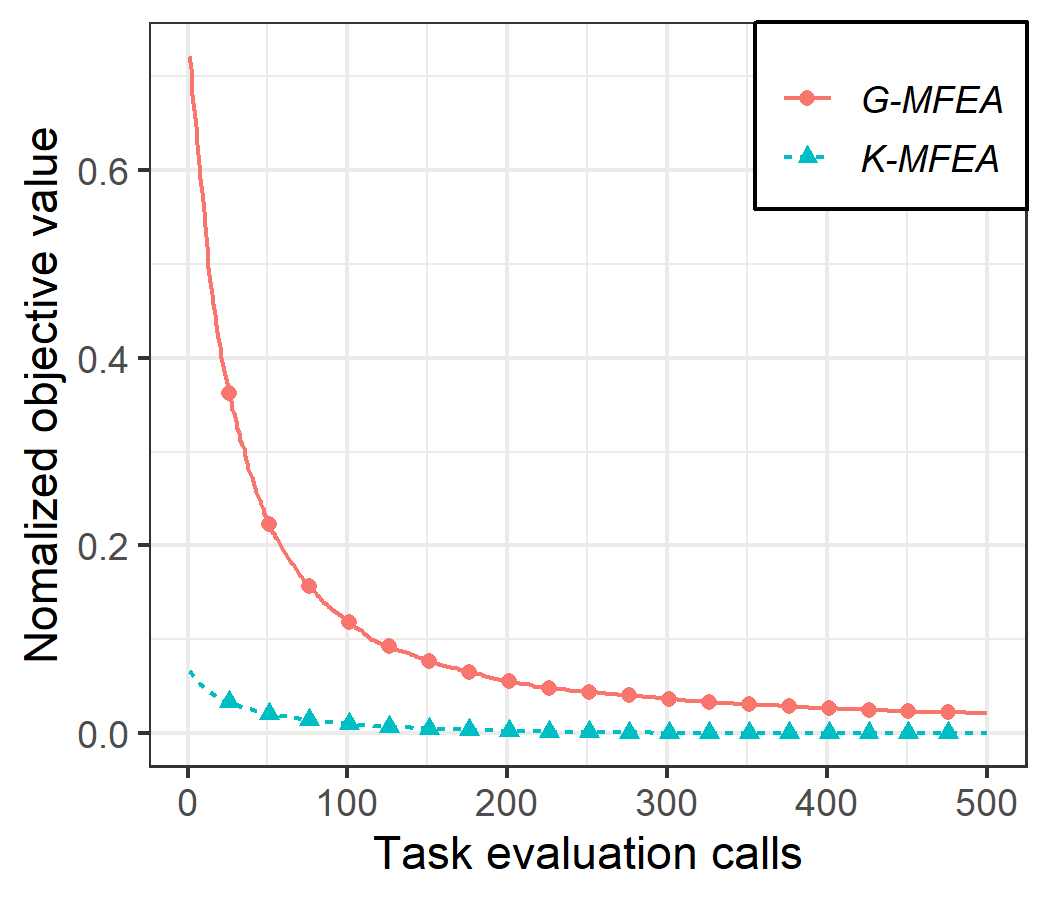}
		\caption{Type 1}
		\label{fig:Type_1_convergence}
	\end{subfigure}
	\begin{subfigure}{.32\linewidth}
		\centering
		\includegraphics[scale=\scalefigure]{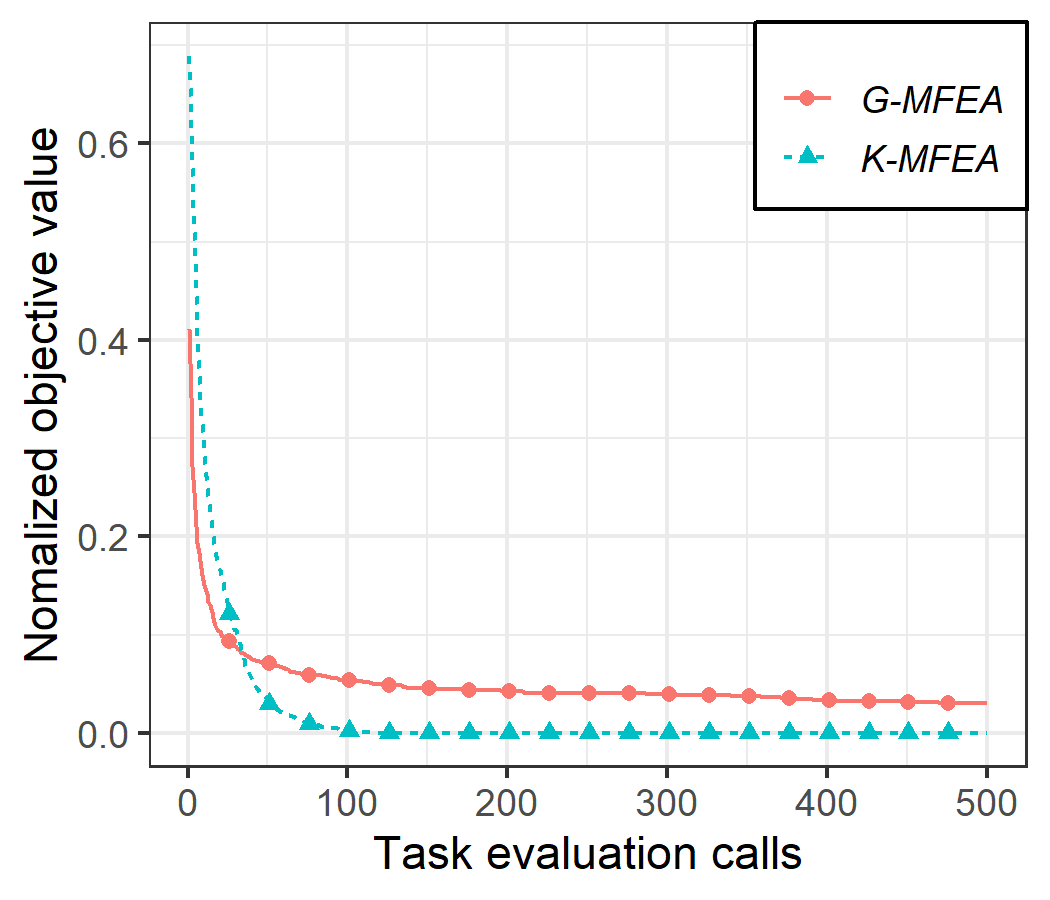}
		\caption{Type 5}
		\label{fig:Type_5_convergence}
	\end{subfigure}
	\begin{subfigure}{.32\linewidth}
		\centering
		\includegraphics[scale=\scalefigure]{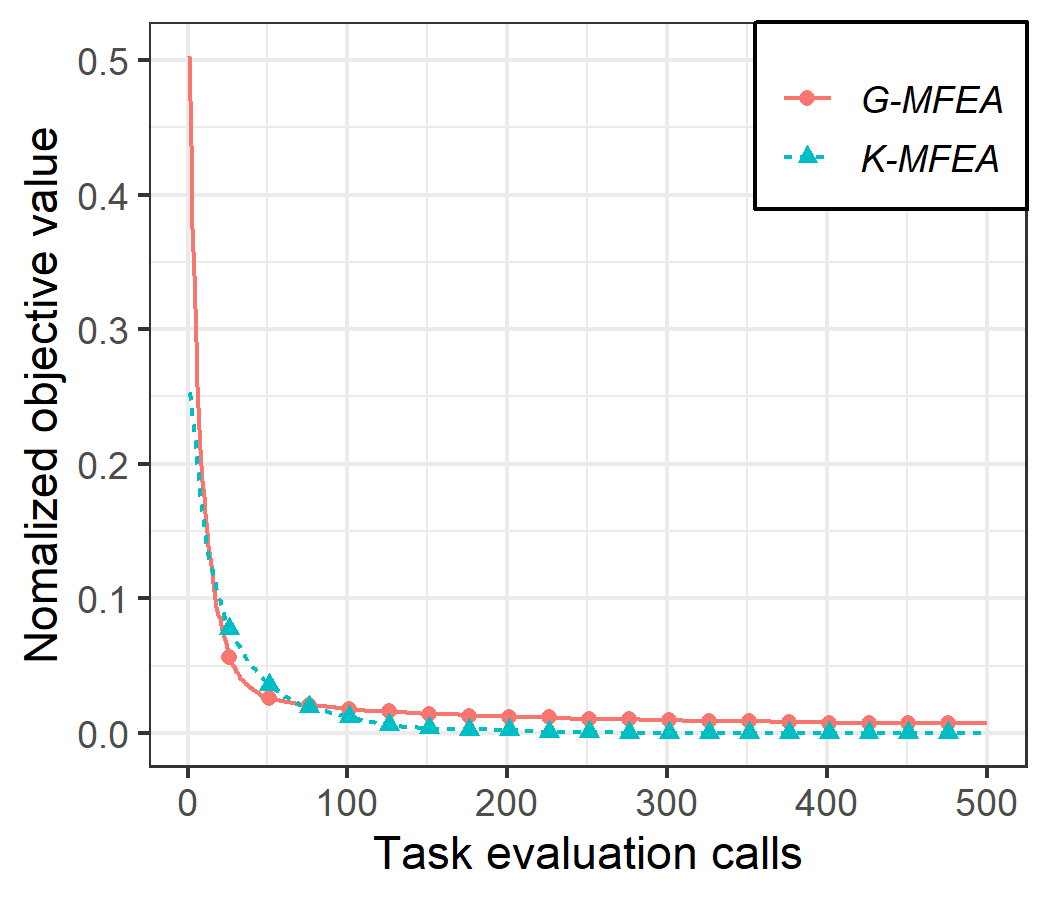}
		\caption{Type 6}
		\label{fig:Type_6_convergence}
	\end{subfigure}
	\caption{Convergence trends of $\tilde{f}$ in multi-tasks and serial single task for instances 10pr439 and 25a280 in Type 1;  instances 5i300-108 and 5i400-205 in Type 5; instances 9pcb442-3x3 and 9pr439-3x3 in Type 6.}
	\label{fig:ConvergenceTrends}
\end{figure}
\setlength{\intextsep}{0pt}

\subsubsection{Analysis influence of the number of parents on results}
\label{subsubsection:AnalysisMultiParents}
In this part, we consider the influence of the number of parents of \gls{mpcx} on the performance of \gls{k-mfea}. We have conducted experiments in the range of 2 to 10 parents, and the results obtained that are different from running with the two parents are shown in Table~\ref{tab:Results_Parents}. In this table, red values indicate better results than values on the left. As shown in the table, the different results are only obtained in large instances and using three or more parents always results in an equal or better experimental result (a better performance is recorded in 12 instances). Besides, the results also show that \gls{mpcx} work most effectively on the parents of 3. 
The more vertices and clusters a graph has, the better result a multi-parent crossover method brings about. Otherwise, using multiple parents in a smaller set is likely unnecessary because a two-parent crossover operator is sufficient to reach the optimal value.
From the above analysis, the \gls{mpcx} is quite suitable in solving large-scale \gls{cluspt} problems.

\subsubsection{Analysis influence of the speed-up evaluation method}
\label{subsubsection:Analysis_SpeedUp_Method}

In this part, we compare the performance of the proposed algorithm on aspect running time when it uses and does not use (called \gls{k-mfea}$^*$) the speed-up method in subsection \ref{subSec:speed_up}.

Table~\ref{tab:CompareSpeedupRunningTime} presents the computational time of algorithms \gls{k-mfea} and \gls{k-mfea}$^*$ with Min (Max) and Avg are minimum (maximum) and average computational time of algorithms on instances of each type, respectively. Results in this table point out that \gls{k-mfea} is faster than \gls{k-mfea}$^*$ on all types of instances. On small types, \gls{k-mfea}'s average running time is 6.9 to 9.9 times smaller. Whereas on large types, \gls{k-mfea} offers 12.8 to 18.8 times less average running time than \gls{k-mfea}$^*$. In particular, \gls{k-mfea}'s average running time is reduced by 54.7 to 80 times compared to \gls{k-mfea}$^*$ on Type 3 and 4 where the average number of vertices per cluster is high.  The method in subsection \ref{subSec:speed_up}, thus, has demonstrated its ability to reduce the computational time of the proposed algorithm.

\setlength{\intextsep}{6pt}
\begin{table}[htbp]
  \centering
  \caption{Running time of the proposed algorithm when uses and does not use the speed-up evaluation method}
  \label{tab:CompareSpeedupRunningTime}
    \begin{tabular}{l rrr rrr}
    	\toprule
          & \multicolumn{3}{c}{\textbf{\glsentrytext{k-mfea}$^*$}} & 
          \multicolumn{3}{c}{\textbf{\glsentrytext{k-mfea}}} \\
          
          \cmidrule(l{3pt}r{3pt}){2-4} 
          \cmidrule(l{3pt}r{3pt}){5-7} 
          
          & \multicolumn{1}{c}{Min} 
          & \multicolumn{1}{c}{Max} 
          & \multicolumn{1}{c}{Avg} 
          
          & \multicolumn{1}{c}{Min} 
          & \multicolumn{1}{c}{Max}
          & \multicolumn{1}{c}{Avg} \\
     \midrule
    Type 1 Small & 0.167 & 0.317 & 0.219 & 0.001 & 0.080 & 0.030 \\
    Type 1 Large & 0.417 & 2.067 & 1.038 & 0.030 & 0.150 & 0.081 \\
     \addlinespace
    Type 3 & 2.533 & 2.867 & 2.720 & 0.030 & 0.050 & 0.034 \\
     \addlinespace
    Type 4 & 1.267 & 2.667 & 1.860 & 0.030 & 0.050 & 0.034 \\
     \addlinespace
    Type 5 Small & 0.117 & 0.300 & 0.208 & 0.020 & 0.030 & 0.021 \\
    Type 5 Large & 0.800 & 2.567 & 1.409 & 0.020 & 0.600 &  0.075\\
     \addlinespace
    Type 6 Small & 0.117 & 0.250 & 0.193 & 0.020 & 0.070 & 0.028 \\
    Type 6 Large & 0.450 & 3.650 & 1.319 & 0.020 & 0.470 & 0.094 \\
    \bottomrule
    \end{tabular}%
  \label{tab:addlabel}%
\end{table}%

\setlength{\intextsep}{6pt}

\section{Conclusion}
\label{Sec_Conclusion}
This paper described a new approach based on \gls{mfea} to solve the \gls{cluspt}. In the proposed algorithm, the solution encoding method is focused on inter-vertices, while the solution evaluation method uses a mechanism that allows reusing the cost of each cluster to be calculated. Therefore, those help to reduce consuming resources on both computation time and memory space. Furthermore, the crossover, mutation, and decoding operators, as well as a method for repairing invalid individuals, are described to enhance the performance of the proposed \gls{mfea}. To evaluate the proposed algorithm, multiple types of Euclidean instances are selected.  The results received from the proposed algorithm are state of the art in most test cases in solving \gls{cluspt} which strongly proven the efficiency of this approach. 

To enhance the performance of the novel algorithm, in the future, the authors will look for evolutionary operators with less complexity.

\section*{Acknowledgements}

This work is funded by Vingroup Joint Stock Company and supported by the Domestic Master/ PhD Scholarship Programme of Vingroup Innovation Foundation (VINIF), Vingroup Big Data Institute (VINBIGDATA) for Ta Bao Thang. This research is funded by the Ministry of Education and Training, Vietnam for the research project: evolutionary multitasking algorithm for the path computation problem in multi-domain networks for Pham Dinh Thanh. This research was sponsored by the U.S. Army Combat Capabilities Development Command (CCDC) Pacific and CCDC Army Research Laboratory (ARL) under Contract Number W90GQZ-93290007 for Huynh Thi Thanh Binh.

\bibliography{references}   

\begin{landscape}
	
\begin{longtable}{l l rrr rrr rrr}
\caption{Results obtained by \glsentrytext{g-mfea}, \glsentrytext{hbrga} and \glsentrytext{k-mfea} on instances in Type 1.}\label{tab:ResultsType1}\\
\toprule
\multicolumn{1}{c}{\textbf{}} & 
\multicolumn{1}{c}{\textbf{}} & 
\multicolumn{3}{c}{\textbf{HB-RGA}} & 
\multicolumn{3}{c}{\textbf{G-MFEA}} & 
\multicolumn{3}{c}{\textbf{K-MFEA}} \\

\cmidrule(l{3pt}r{3pt}){3-5} 
\cmidrule(l{3pt}r{3pt}){6-8} 
\cmidrule(l{3pt}r{3pt}){9-11}

 & Instances & 
BF & 
Avg & 
Time & 
BF & 
Avg & 
Time & 
BF & 
Avg & 
Time
\\
\midrule
 & 10berlin52 & 43738.6 & \textcolor{black}{43971} & 0.04 & 43724.1 & \textcolor{black}{43724.1} & 0.67 & 43724.1 & \textcolor{black}{43724.1} & 0.02\\
 
 & 10eil51 & 1713.2 & \textcolor{black}{1723.2} & 0.04 & 1713.2 & \textcolor{black}{1713.2} & 0.92 & 1713.2 & \textcolor{black}{1713.2} & 0.02\\
 
 & 10eil76 & 2203.3 & \textcolor{black}{2208.4} & 0.05 & 2203.3 & \textcolor{black}{2203.3} & 1.60 & 2203.3 & \textcolor{black}{2203.3} & 0.02\\
 
 & 10kroB100 & 140635.1 & \textcolor{black}{141951.4} & 0.06 & 140551.2 & \textcolor{black}{140597.9} & 2.35 & 140522.2 & \textcolor{red}{\em{140522.2}} & 0.02\\
 
 & 10pr76 & 522572.2 & \textcolor{black}{525733.1} & 0.04 & 522213.8 & \textcolor{black}{522340.4} & 1.67 & 522213.8 & \textcolor{red}{\em{522213.8}} & 0.02\\
 \addlinespace
 & 10rat99 & 7520.2 & \textcolor{black}{7562.1} & 0.05 & 7520.2 & \textcolor{black}{7524} & 2.35 & 7520.2 & \textcolor{red}{\em{7520.2}} & 0.02\\
 
 & 10st70 & 3099.5 & \textcolor{black}{3131.7} & 0.04 & 3095.2 & \textcolor{black}{3095.7} & 1.40 & 3095.2 & \textcolor{red}{\em{3095.2}} & 0.02\\
 
 & 15berlin52 & 26312.0 & \textcolor{black}{26437.7} & 0.03 & 26315.5 & \textcolor{black}{26351.7} & 1.07 & 26312.0 & \textcolor{red}{\em{26312}} & 0.02\\
 
 & 15eil51 & 1306.7 & \textcolor{black}{1313.8} & 0.03 & 1306.8 & \textcolor{black}{1309.1} & 1.00 & 1306.4 & \textcolor{red}{\em{1306.4}} & 0.02\\
 
 & 15eil76 & 2911.3 & \textcolor{black}{2921.8} & 0.04 & 2909.1 & \textcolor{black}{2913.1} & 1.62 & 2909.1 & \textcolor{red}{\em{2909.1}} & 0.02\\
 \addlinespace
 & 15pr76 & 705017.3 & \textcolor{black}{708944.9} & 0.04 & 705226.1 & \textcolor{black}{706505.5} & 1.63 & 704600.6 & \textcolor{red}{\em{704600.6}} & 0.02\\
 
 & 15st70 & 4129.9 & \textcolor{black}{4147.3} & 0.03 & 4126.7 & \textcolor{black}{4135.5} & 1.40 & 4120.1 & \textcolor{red}{\em{4120.1}} & 0.02\\
 
 & 25eil101 & 4680.8 & \textcolor{black}{4686.1} & 0.05 & 4700.4 & \textcolor{black}{4727.9} & 2.38 & 4679.0 & \textcolor{red}{\em{4679}} & 0.03\\
 
 & 25kroA100 & 147239.0 & \textcolor{black}{147716.8} & 0.05 & 148767.9 & \textcolor{black}{149708.1} & 2.37 & 147195.0 & \textcolor{red}{\em{147195}} & 0.03\\
 
 & 25lin105 & 98087.5 & \textcolor{black}{98502.9} & 0.05 & 98941.4 & \textcolor{black}{100585.3} & 2.50 & 97944.7 & \textcolor{red}{\em{97944.7}} & 0.03\\
 \addlinespace
 & 25rat99 & 6846.3 & \textcolor{black}{6867.8} & 0.05 & 6930.9 & \textcolor{black}{7022.3} & 2.33 & 6841.5 & \textcolor{red}{\em{6841.5}} & 0.03\\
 
 & 50eil101 & 3827.3 & \textcolor{black}{3828.1} & 0.05 & 4034.7 & \textcolor{black}{4178.1} & 3.15 & 3825.3 & \textcolor{red}{\em{3825.3}} & 0.08\\
 
 & 50kroA100 & 159815.2 & \textcolor{black}{160029.9} & 0.05 & 173113.3 & \textcolor{black}{179506.1} & 3.13 & 159647.2 & \textcolor{red}{\em{159647.2}} & 0.08\\
 
 & 50kroB100 & 133135.4 & \textcolor{black}{133325.8} & 0.06 & 149465.6 & \textcolor{black}{157831.1} & 3.13 & 133104.5 & \textcolor{red}{\em{133104.5}} & 0.08\\
 
 & 50lin105 & 145869.9 & \textcolor{black}{145951.8} & 0.06 & 151901.5 & \textcolor{black}{154680.7} & 3.33 & 145829.1 & \textcolor{red}{\em{145829.1}} & 0.08\\
 \addlinespace
 & 50rat99 & 8010.6 & \textcolor{black}{8016.8} & 0.05 & 8728.0 & \textcolor{black}{9002.1} & 3.15 & 8007.4 & \textcolor{red}{\em{8007.4}} & 0.05\\
 
 & 5berlin52 & 22746.4 & \textcolor{black}{23106.9} & 0.03 & 22746.4 & \textcolor{black}{22746.4} & 1.02 & 22746.4 & \textcolor{black}{22746.4} & 0.05\\
 
 & 5eil51 & 1770.5 & \textcolor{black}{1792.3} & 0.03 & 1769.4 & \textcolor{black}{1769.4} & 0.95 & 1769.4 & \textcolor{black}{1769.4} & 0.00\\
 
 & 5eil76 & 2630.8 & \textcolor{black}{2658.4} & 0.05 & 2630.8 & \textcolor{black}{2630.8} & 1.83 & 2630.8 & \textcolor{black}{2630.8} & 0.00\\
 
 & 5pr76 & 585008.0 & \textcolor{black}{589778.1} & 0.06 & 585008.0 & \textcolor{black}{585008} & 1.82 & 585008.0 & \textcolor{black}{585008} & 0.00\\
 \addlinespace
 
\multirow{-26}{*}{\centering \begin{sideways} \textbf{Small instances} \end{sideways}} 

& 5st70 & 4520.1 & \textcolor{black}{4562.8} & 0.04 & 4520.1 & \textcolor{black}{4520.1} & 1.57 & 4520.1 & \textcolor{black}{4520.1} & 0.00\\

\cmidrule{1-11}

 & 10a280 & 28193.6 & \textcolor{black}{28515.8} & 0.31 & 27936.1 & \textcolor{black}{28079.4} & 14.72 & 27925.2 & \textcolor{red}{\em{27925.2}} & 0.03\\
 
 & 10gil262 & 27653.6 & \textcolor{black}{27788} & 0.27 & 27637.5 & \textcolor{black}{27645.7} & 12.78 & 27637.5 & \textcolor{red}{\em{27637.5}} & 0.03\\
 
 & 10lin318 & 815294.7 & \textcolor{black}{825808.2} & 0.37 & 812744.1 & \textcolor{black}{814264.5} & 18.55 & 809750.0 & \textcolor{red}{\em{809750}} & 0.03\\
 
 & 10pcb442 & 745264.4 & \textcolor{black}{750362.1} & 0.87 & 742112.4 & \textcolor{black}{742678.9} & 39.38 & 741195.8 & \textcolor{red}{\em{741195.8}} & 0.03\\
 
 & 10pr439 & 1917984.5 & \textcolor{black}{1930766.4} & 1.32 & 1907568.9 & \textcolor{black}{1911815.3} & 42.50 & 1904690.2 & \textcolor{red}{\em{1904718.9}} & 0.05\\
 \addlinespace
 & 25a280 & 30058.6 & \textcolor{black}{30228.5} & 0.23 & 30373.2 & \textcolor{black}{30654.1} & 11.17 & 29902.4 & \textcolor{red}{\em{29902.4}} & 0.05\\
 
 & 25gil262 & 30535.7 & \textcolor{black}{30711.1} & 0.22 & 30695.8 & \textcolor{black}{30953.2} & 11.67 & 30325.7 & \textcolor{red}{\em{30325.7}} & 0.07\\
 
 & 25lin318 & 592166.4 & \textcolor{black}{595489} & 0.31 & 593017.4 & \textcolor{black}{601118.8} & 16.65 & 584554.0 & \textcolor{red}{\em{584554}} & 0.07\\
 
 & 25pcb442 & 744734.7 & \textcolor{black}{748031.6} & 0.54 & 757524.1 & \textcolor{black}{762707.5} & 31.28 & 740892.6 & \textcolor{red}{\em{740892.6}} & 0.07\\
 
 & 25pr439 & 1519322.1 & \textcolor{black}{1531976.6} & 0.89 & 1531948.9 & \textcolor{black}{1556475.7} & 29.62 & 1511168.9 & \textcolor{red}{\em{1511197.2}} & 0.07\\
 \addlinespace
 & 50a280 & 36406.5 & \textcolor{black}{36471} & 0.22 & 38596.8 & \textcolor{black}{39872.5} & 12.15 & 36266.9 & \textcolor{red}{\em{36266.9}} & 0.13\\
 
 & 50gil262 & 26634.0 & \textcolor{black}{26680.8} & 0.22 & 28780.1 & \textcolor{black}{30004} & 12.12 & 26523.3 & \textcolor{red}{\em{26523.3}} & 0.13\\
 
 & 50lin318 & 690743.8 & \textcolor{black}{691897.5} & 0.31 & 730023.7 & \textcolor{black}{748203.2} & 16.80 & 688724.6 & \textcolor{red}{\em{688748.6}} & 0.15\\
 
 & 50pcb442 & 913395.0 & \textcolor{black}{914652.8} & 0.57 & 962101.7 & \textcolor{black}{990436} & 31.00 & 910478.7 & \textcolor{red}{\em{910487.5}} & 0.15\\
 
\multirow{-15}{*}{\centering \begin{sideways} \textbf{Large instances} \end{sideways}} 

& 50pr439 & 2162274.5 & \textcolor{black}{2167713.8} & 0.93 & 2302262.5 & \textcolor{black}{2375177.6} & 25.13 & 2152986.6 & \textcolor{red}{\em{2152987.9}} & 0.15\\
\bottomrule
\end{longtable}	

\begin{longtable}{l l rrr rrr rrr}
	\caption{Results obtained by \glsentrytext{g-mfea}, \glsentrytext{hbrga} and \glsentrytext{k-mfea} on instances in Types 3 and 4.}\label{tab:ResultsType34}\\
	\toprule
	\multicolumn{1}{c}{\textbf{}} & 
	\multicolumn{1}{c}{\textbf{}} & 
	\multicolumn{3}{c}{\textbf{HB-RGA}} & 
	\multicolumn{3}{c}{\textbf{G-MFEA}} & 
	\multicolumn{3}{c}{\textbf{K-MFEA}} \\

	\cmidrule(l{3pt}r{3pt}){3-5} 
	\cmidrule(l{3pt}r{3pt}){6-8} 
	\cmidrule(l{3pt}r{3pt}){9-11}
	
	& Instances & 
	BF & 
	Avg & 
	Time & 
	BF & 
	Avg & 
	Time & 
	BF & 
	Avg & 
	Time
	\\
\midrule
 & 6i300 & 19461.1 & \textcolor{black}{19836.6} & 0.55 & 19286.3 & \textcolor{black}{19320.3} & 23.42 & 19264.5 & \textcolor{red}{\em{19264.5}} & 0.03\\
 
 & 6i350 & 21385.6 & \textcolor{black}{21713.4} & 0.73 & 21218.5 & \textcolor{black}{21261.3} & 33.20 & 21217.2 & \textcolor{red}{\em{21217.2}} & 0.03\\
 
 & 6i400 & 29513.4 & \textcolor{black}{29913.7} & 1.05 & 29389.5 & \textcolor{black}{29437.5} & 46.60 & 29348.2 & \textcolor{red}{\em{29348.2}} & 0.03\\
 
 & 6i450 & 35899.2 & \textcolor{black}{36463.5} & 1.34 & 35715.7 & \textcolor{black}{35795.9} & 60.45 & 35681.5 & \textcolor{red}{\em{35681.5}} & 0.03\\
 
\multirow{-5}{*}{\raggedright\arraybackslash Type 3} 

& 6i500 & 37799.5 & \textcolor{black}{38225.9} & 1.95 & 37567.6 & \textcolor{black}{37631.4} & 82.30 & 37510.1 & \textcolor{red}{\em{37516.1}} & 0.05\\

\cmidrule{1-11}

 & 4i200a & 97959.6 & \textcolor{black}{97974.1} & 0.35 & 97959.6 & \textcolor{black}{97959.6} & 11.43 & 97959.6 & \textcolor{black}{97959.6} & 0.02\\
 
 & 4i200h & 87675.3 & \textcolor{black}{88285.1} & 0.36 & 87675.3 & \textcolor{black}{87675.3} & 11.63 & 87675.3 & \textcolor{black}{87675.3} & 0.02\\
 
 & 4i200x1 & 123811.4 & \textcolor{black}{124825.9} & 0.35 & 123669.7 & \textcolor{black}{123670.2} & 11.67 & 123669.7 & \textcolor{red}{\em{123669.7}} & 0.02\\
 
 & 4i200x2 & 114059.5 & \textcolor{black}{115432} & 0.37 & 114012.3 & \textcolor{black}{114012.3} & 12.03 & 114012.3 & \textcolor{black}{114012.3} & 0.02\\
 
 & 4i200z & 131807.4 & \textcolor{black}{133697.8} & 0.38 & 131683.5 & \textcolor{black}{131685.2} & 11.57 & 131683.5 & \textcolor{red}{\em{131683.5}} & 0.02\\
  \addlinespace
 & 4i400a & 214115.3 & \textcolor{black}{214230.5} & 2.53 & 214115.3 & \textcolor{black}{214115.3} & 64.53 & 214115.3 & \textcolor{black}{214115.3} & 0.02\\
 
 & 4i400h & 257054.0 & \textcolor{black}{260183.1} & 2.41 & 256200.5 & \textcolor{black}{256291.2} & 62.68 & 256200.5 & \textcolor{red}{\em{256200.5}} & 0.03\\
 
 & 4i400x1 & 188840.1 & \textcolor{black}{191145.3} & 2.32 & 199389.3 & \textcolor{black}{222805.4} & 36.93 & 188196.7 & \textcolor{red}{\em{188196.7}} & 0.03\\
 
 & 4i400x2 & 159254.8 & \textcolor{black}{162685.6} & 2.35 & 176188.2 & \textcolor{black}{195580.8} & 64.03 & 159254.8 & \textcolor{red}{\em{159254.8}} & 0.03\\
 
\multirow{-10}{*}{\raggedright\arraybackslash Type 4}

 & 4i400z & 221460.6 & \textcolor{black}{224677.4} & 2.22 & 234203.3 & \textcolor{black}{244894.6} & 36.95 & 221423.9 & \textcolor{red}{\em{221423.9}} & 0.03\\
\bottomrule
\end{longtable}

\begin{longtable}{l l rrr rrr rrr}
	\caption{Results obtained by \glsentrytext{g-mfea}, \glsentrytext{hbrga} and \glsentrytext{k-mfea} on instances in Type 5.}\label{tab:ResultsType5}\\
	\toprule
	\multicolumn{1}{c}{\textbf{}} & 
	\multicolumn{1}{c}{\textbf{}} & 
	\multicolumn{3}{c}{\textbf{HB-RGA}} & 
	\multicolumn{3}{c}{\textbf{G-MFEA}} & 
	\multicolumn{3}{c}{\textbf{K-MFEA}} \\

	\cmidrule(l{3pt}r{3pt}){3-5} 
	\cmidrule(l{3pt}r{3pt}){6-8} 
	\cmidrule(l{3pt}r{3pt}){9-11}
	
	& Instances & 
	BF & 
	Avg & 
	Time & 
	BF & 
	Avg & 
	Time & 
	BF & 
	Avg & 
	Time
	\\
\midrule
 & 10i120-46 & 94055.2 & \textcolor{black}{94596.7} & 0.08 & 93956.9 & \textcolor{black}{94034.3} & 2.77 & 93925.0 & \textcolor{red}{\em{93925}} & 0.02\\
 
& 10i30-17 & 13276.6 & \textcolor{black}{13289.2} & 0.02 & 13276.6 & \textcolor{black}{13276.6} & 0.58 & 13276.6 & \textcolor{black}{13276.6} & 0.02\\
 
& 10i45-18 & 23267.6 & \textcolor{black}{23344.8} & 0.03 & 22890.4 & \textcolor{black}{22892.2} & 0.85 & 22890.4 & \textcolor{red}{\em{22890.4}} & 0.02\\
 
& 10i60-21 & 33744.5 & \textcolor{black}{35002} & 0.03 & 33694.8 & \textcolor{black}{33702.8} & 1.20 & 33694.8 & \textcolor{red}{\em{33694.8}} & 0.02\\
 
& 10i65-21 & 37386.7 & \textcolor{black}{37677} & 0.04 & 37353.1 & \textcolor{black}{37353.6} & 1.32 & 37353.1 & \textcolor{red}{\em{37353.1}} & 0.02\\
 \addlinespace
& 10i70-21 & 38543.8 & \textcolor{black}{38855} & 0.03 & 38066.7 & \textcolor{black}{38187.3} & 1.43 & 38059.5 & \textcolor{red}{\em{38059.5}} & 0.02\\
 
& 10i75-22 & 65411.9 & \textcolor{black}{65783.1} & 0.05 & 65362.0 & \textcolor{black}{65397.3} & 1.67 & 65361.9 & \textcolor{red}{\em{65361.9}} & 0.03\\
 
& 10i90-33 & 52091.2 & \textcolor{black}{52617.6} & 0.05 & 51943.2 & \textcolor{black}{51975.6} & 2.03 & 51931.2 & \textcolor{red}{\em{51931.2}} & 0.03\\
 
& 5i120-46 & 61776.0 & \textcolor{black}{62393.2} & 0.11 & 61451.5 & \textcolor{black}{61495.3} & 3.65 & 61451.5 & \textcolor{red}{\em{61451.5}} & 0.02\\
 
& 5i30-17 & 14399.9 & \textcolor{black}{14399.9} & 0.02 & 14399.9 & \textcolor{black}{14399.9} & 0.62 & 14399.9 & \textcolor{black}{14399.9} & 0.02\\
 \addlinespace
& 5i45-18 & 14884.3 & \textcolor{black}{14893} & 0.02 & 14884.3 & \textcolor{black}{14884.3} & 0.88 & 14884.3 & \textcolor{black}{14884.3} & 0.02\\
 
& 5i60-21 & 28422.7 & \textcolor{black}{28584.2} & 0.03 & 28422.7 & \textcolor{black}{28422.7} & 1.18 & 28422.7 & \textcolor{black}{28422.7} & 0.02\\
 
& 5i65-21 & 31244.3 & \textcolor{black}{31684.7} & 0.03 & 30907.8 & \textcolor{black}{30911.7} & 1.40 & 30907.8 & \textcolor{red}{\em{30907.8}} & 0.02\\
 
& 5i70-21 & 35052.8 & \textcolor{black}{35384.4} & 0.04 & 35052.8 & \textcolor{black}{35052.8} & 1.57 & 35052.8 & \textcolor{black}{35052.8} & 0.02\\
 
& 5i75-22 & 34811.1 & \textcolor{black}{34993.8} & 0.05 & 34692.5 & \textcolor{black}{34692.5} & 1.83 & 34692.5 & \textcolor{black}{34692.5} & 0.02\\
 \addlinespace
& 5i90-33 & 52128.9 & \textcolor{black}{52916} & 0.06 & 51977.0 & \textcolor{black}{51977.3} & 2.28 & 51977.0 & \textcolor{red}{\em{51977}} & 0.02\\
 
& 7i30-17 & 20438.9 & \textcolor{black}{20450} & 0.02 & 20438.9 & \textcolor{black}{20438.9} & 0.57 & 20438.9 & \textcolor{black}{20438.9} & 0.02\\
 
& 7i45-18 & 20512.0 & \textcolor{black}{20973.8} & 0.02 & 20512.0 & \textcolor{black}{20512} & 0.80 & 20512.0 & \textcolor{black}{20512} & 0.02\\
 
& 7i60-21 & 36263.9 & \textcolor{black}{36339.5} & 0.03 & 36263.9 & \textcolor{black}{36263.9} & 1.13 & 36263.9 & \textcolor{black}{36263.9} & 0.02\\
 
\multirow{-20}{*}{\centering \begin{sideways} \textbf{Small instances} \end{sideways}} 

 & 7i65-21 & 34847.6 & \textcolor{black}{34881.9} & 0.04 & 34847.6 & \textcolor{black}{34847.6} & 1.33 & 34847.6 & \textcolor{black}{34847.6} & 0.02\\
 
\cmidrule{1-11}

& 10i300-109 & 113292.8 & \textcolor{black}{114274.1} & 0.33 & 112876.2 & \textcolor{black}{113017.1} & 16.98 & 112681.0 & \textcolor{red}{\em{112681}} & 0.03\\
 
& 10i400-206 & 209409.3 & \textcolor{black}{211253.4} & 0.60 & 207778.5 & \textcolor{black}{208087.4} & 31.85 & 207521.7 & \textcolor{red}{\em{207521.7}} & 0.03\\
 
& 10i500-305 & 352151.4 & \textcolor{black}{356961.1} & 0.99 & 350897.4 & \textcolor{black}{351929.6} & 52.50 & 349675.2 & \textcolor{red}{\em{349675.2}} & 0.03\\
 
& 15i300-110 & 114092.2 & \textcolor{black}{114531.8} & 0.28 & 112935.9 & \textcolor{black}{113358.9} & 15.83 & 112096.7 & \textcolor{red}{\em{112096.7}} & 0.03\\
 
& 15i400-207 & 165565.7 & \textcolor{black}{166796.4} & 0.47 & 165328.2 & \textcolor{black}{165854.9} & 28.27 & 164117.8 & \textcolor{red}{\em{164117.8}} & 0.03\\
 \addlinespace
& 15i500-306 & 305034.5 & \textcolor{black}{306929.7} & 0.80 & 304128.7 & \textcolor{black}{304949.4} & 42.77 & 300734.1 & \textcolor{red}{\em{300734.1}} & 0.03\\
 
& 20i300-111 & 157526.1 & \textcolor{black}{158563.7} & 0.24 & 157371.4 & \textcolor{black}{157990.3} & 15.30 & 156347.7 & \textcolor{red}{\em{156347.7}} & 0.05\\
 
& 20i400-208 & 226173.0 & \textcolor{black}{226963.7} & 0.39 & 226383.9 & \textcolor{black}{227069.5} & 26.78 & 224012.5 & \textcolor{red}{\em{224012.5}} & 0.05\\
 
& 20i500-307 & 203339.4 & \textcolor{black}{204679.4} & 0.64 & 202938.5 & \textcolor{black}{203910.7} & 39.17 & 200328.7 & \textcolor{red}{\em{200343}} & 0.05\\
 
& 25i300-112 & 117661.9 & \textcolor{black}{118121.6} & 0.22 & 118310.1 & \textcolor{black}{119401.3} & 14.38 & 116193.6 & \textcolor{red}{\em{116193.6}} & 0.05\\
 \addlinespace
& 25i400-209 & 231886.7 & \textcolor{black}{233015.6} & 0.38 & 233552.7 & \textcolor{black}{236456.6} & 23.27 & 229913.6 & \textcolor{red}{\em{229913.6}} & 0.05\\
 
& 25i500-308 & 300838.0 & \textcolor{black}{301906.8} & 0.61 & 302497.1 & \textcolor{black}{304046.5} & 24.60 & 299498.2 & \textcolor{red}{\em{299498.2}} & 0.05\\
 
& 5i300-108 & 177698.0 & \textcolor{black}{179796.8} & 0.67 & 177185.9 & \textcolor{black}{177220.7} & 26.78 & 177185.9 & \textcolor{red}{\em{177185.9}} & 0.02\\
 
& 5i400-205 & 210229.5 & \textcolor{black}{216014.7} & 1.17 & 209488.0 & \textcolor{black}{209970.9} & 52.22 & 209389.8 & \textcolor{red}{\em{209389.8}} & 0.02\\
 
\multirow{-15}{*}{\centering \begin{sideways} \textbf{Large instances} \end{sideways}} 

& 5i500-304 & 183130.1 & \textcolor{black}{185809.5} & 2.11 & 182206.2 & \textcolor{red}{\em{182416.4}} & 90.62 & 182357.0 & \textcolor{black}{183976.6} & 0.60\\
\bottomrule
\end{longtable}

\begin{longtable}{l l rrr rrr rrr}
	\caption{Results obtained by \glsentrytext{g-mfea}, \glsentrytext{hbrga} and \glsentrytext{k-mfea} on instances in Type 6.}\label{tab:ResultsType6}\\
	\toprule
	\multicolumn{1}{c}{\textbf{}} & 
	\multicolumn{1}{c}{\textbf{}} & 
	\multicolumn{3}{c}{\textbf{HB-RGA}} & 
	\multicolumn{3}{c}{\textbf{G-MFEA}} & 
	\multicolumn{3}{c}{\textbf{K-MFEA}} \\

	\cmidrule(l{3pt}r{3pt}){3-5} 
	\cmidrule(l{3pt}r{3pt}){6-8} 
	\cmidrule(l{3pt}r{3pt}){9-11}
	
	& Instances & 
	BF & 
	Avg & 
	Time & 
	BF & 
	Avg & 
	Time & 
	BF & 
	Avg & 
	Time
	\\
\midrule
 & 10berlin52-2x5 & 27472.4 & \textcolor{black}{27723.2} & 0.04 & 27471.4 & \textcolor{black}{27473} & 0.83 & 27471.4 & \textcolor{red}{\em{27471.4}} & 0.02\\
 
 & 12eil51-3x4 & 1699.1 & \textcolor{black}{1702.3} & 0.03 & 1699.0 & \textcolor{black}{1699.1} & 0.93 & 1699.0 & \textcolor{red}{\em{1699}} & 0.02\\
 
 & 12eil76-3x4 & 2650.8 & \textcolor{black}{2653.2} & 0.04 & 2650.8 & \textcolor{black}{2650.8} & 1.57 & 2650.8 & \textcolor{black}{2650.8} & 0.03\\
 
 & 12pr76-3x4 & 600597.6 & \textcolor{black}{603474.4} & 0.05 & 600430.9 & \textcolor{black}{600818.7} & 1.65 & 600008.6 & \textcolor{red}{\em{600008.6}} & 0.03\\
 
 & 12st70-3x4 & 4128.1 & \textcolor{black}{4144.6} & 0.03 & 4106.5 & \textcolor{black}{4110.1} & 1.43 & 4106.5 & \textcolor{red}{\em{4106.5}} & 0.03\\
  \addlinespace
 & 15pr76-3x5 & 526596.7 & \textcolor{black}{532896.8} & 0.06 & 525170.3 & \textcolor{black}{526166.2} & 1.63 & 524335.2 & \textcolor{red}{\em{524335.2}} & 0.03\\
  
 & 16eil51-4x4 & 1302.4 & \textcolor{black}{1304} & 0.02 & 1302.7 & \textcolor{black}{1305.6} & 1.00 & 1301.4 & \textcolor{red}{\em{1301.4}} & 0.03\\
 
 & 16eil76-4x4 & 2042.4 & \textcolor{black}{2053.8} & 0.04 & 2040.0 & \textcolor{black}{2052.2} & 1.67 & 2036.0 & \textcolor{red}{\em{2036}} & 0.03\\
 
 & 16lin105-4x4 & 125052.2 & \textcolor{black}{125685.3} & 0.06 & 125052.2 & \textcolor{black}{125289.8} & 2.40 & 125052.2 & \textcolor{red}{\em{125052.2}} & 0.03\\
 
 & 16st70-4x4 & 2939.3 & \textcolor{black}{2966.1} & 0.03 & 2935.4 & \textcolor{black}{2949.2} & 1.45 & 2932.6 & \textcolor{red}{\em{2932.6}} & 0.03\\
  \addlinespace
 & 18pr76-3x6 & 642733.0 & \textcolor{black}{646399.1} & 0.04 & 639723.3 & \textcolor{black}{641700.1} & 1.68 & 638164.5 & \textcolor{red}{\em{638164.5}} & 0.03\\
 
 & 20eil51-4x5 & 2284.2 & \textcolor{black}{2287.9} & 0.03 & 2288.7 & \textcolor{black}{2295.2} & 1.05 & 2283.7 & \textcolor{red}{\em{2283.7}} & 0.03\\
  
 & 20eil76-4x5 & 2385.9 & \textcolor{black}{2392.6} & 0.03 & 2390.5 & \textcolor{black}{2402.2} & 1.68 & 2385.9 & \textcolor{red}{\em{2385.9}} & 0.03\\
 
 & 20st70-4x5 & 2939.4 & \textcolor{black}{2945.7} & 0.03 & 2942.8 & \textcolor{black}{2967} & 1.47 & 2934.8 & \textcolor{red}{\em{2934.8}} & 0.03\\
 
 & 25eil101-5x5 & 3609.1 & \textcolor{black}{3622} & 0.05 & 3649.2 & \textcolor{black}{3670.5} & 2.50 & 3603.5 & \textcolor{red}{\em{3603.5}} & 0.03\\
  \addlinespace
 & 25eil51-5x5 & 1474.6 & \textcolor{black}{1476.2} & 0.02 & 1487.4 & \textcolor{black}{1507.3} & 1.18 & 1474.6 & \textcolor{red}{\em{1474.6}} & 0.03\\
 
 & 25eil76-5x5 & 2193.1 & \textcolor{black}{2194.8} & 0.03 & 2219.1 & \textcolor{black}{2245.2} & 1.73 & 2193.1 & \textcolor{red}{\em{2193.1}} & 0.03\\
 
 & 25rat99-5x5 & 11400.3 & \textcolor{black}{11418.9} & 0.05 & 11434.9 & \textcolor{black}{11485.9} & 2.33 & 11395.8 & \textcolor{red}{\em{11395.8}} & 0.03\\
 
 & 28kroA100-4x7 & 134129.0 & \textcolor{black}{134532.3} & 0.05 & 136501.1 & \textcolor{black}{138342.8} & 2.38 & 133101.6 & \textcolor{red}{\em{133101.6}} & 0.03\\
 
 & 30kroB100-5x6 & 198976.7 & \textcolor{black}{199205.7} & 0.05 & 200596.8 & \textcolor{black}{202209.8} & 2.50 & 197934.6 & \textcolor{red}{\em{197934.6}} & 0.03\\
  \addlinespace
 & 35kroB100-5x5 & 129122.6 & \textcolor{black}{129832.3} & 0.05 & 130935.1 & \textcolor{black}{132840.4} & 2.48 & 129078.7 & \textcolor{red}{\em{129078.7}} & 0.03\\
 
 & 36eil101-6x6 & 3850.7 & \textcolor{black}{3852.1} & 0.04 & 3929.2 & \textcolor{black}{3981.6} & 2.68 & 3850.7 & \textcolor{red}{\em{3850.7}} & 0.07\\
 
 & 42rat99-6x7 & 8902.5 & \textcolor{black}{8906} & 0.05 & 9187.0 & \textcolor{black}{9393.5} & 2.75 & 8902.1 & \textcolor{red}{\em{8902.1}} & 0.07\\
 
 & 4berlin52-2x2 & 23287.9 & \textcolor{black}{23395.9} & 0.03 & 23287.9 & \textcolor{black}{23287.9} & 1.10 & 23287.9 & \textcolor{black}{23287.9} & 0.02\\
 
 & 4eil51-2x2 & 1898.5 & \textcolor{black}{1915.6} & 0.03 & 1898.5 & \textcolor{black}{1898.5} & 0.97 & 1898.5 & \textcolor{black}{1898.5} & 0.02\\
  \addlinespace
 & 4eil76-2x2 & 2948.8 & \textcolor{black}{2974.6} & 0.05 & 2948.7 & \textcolor{black}{2948.7} & 1.98 & 2948.7 & \textcolor{black}{2948.7} & 0.02\\
 
 & 4pr76-2x2 & 442693.0 & \textcolor{black}{445997.3} & 0.06 & 442693.0 & \textcolor{black}{442693} & 2.02 & 442693.0 & \textcolor{black}{442693} & 0.02\\
 
 & 6berlin52-2x3 & 32130.8 & \textcolor{black}{32295.7} & 0.03 & 32128.6 & \textcolor{black}{32128.6} & 1.13 & 32128.6 & \textcolor{black}{32128.6} & 0.02\\
 
 & 6pr76-2x3 & 648884.9 & \textcolor{black}{656228.6} & 0.05 & 648275.7 & \textcolor{black}{648507.9} & 1.73 & 648275.7 & \textcolor{red}{\em{648275.7}} & 0.02\\
 
 & 6st70-2x3 & 3476.7 & \textcolor{black}{3503.5} & 0.04 & 3476.7 & \textcolor{black}{3476.7} & 1.40 & 3476.7 & \textcolor{black}{3476.7} & 0.02\\
  \addlinespace
 & 8berlin52-2x4 & 26854.4 & \textcolor{black}{26969.8} & 0.03 & 26783.2 & \textcolor{black}{26795.4} & 0.93 & 26783.2 & \textcolor{red}{\em{26783.2}} & 0.02\\
 
 & 9eil101-3x3 & 3135.4 & \textcolor{black}{3154.3} & 0.06 & 3117.6 & \textcolor{black}{3120.2} & 2.32 & 3117.6 & \textcolor{red}{\em{3117.6}} & 0.02\\
 
 & 9eil51-3x3 & 1912.8 & \textcolor{black}{1921.3} & 0.03 & 1908.0 & \textcolor{black}{1909.9} & 0.92 & 1907.7 & \textcolor{red}{\em{1907.7}} & 0.02\\
 
 & 9eil76-3x3 & 2938.4 & \textcolor{black}{2956.6} & 0.04 & 2937.4 & \textcolor{black}{2938.6} & 1.57 & 2937.4 & \textcolor{red}{\em{2937.4}} & 0.02\\
 
\multirow{-35}{*}{\centering \begin{sideways} \textbf{Small instances} \end{sideways}} 

& 9pr76-3x3 & 554995.8 & \textcolor{black}{560230.3} & 0.04 & 553685.6 & \textcolor{black}{553849.7} & 1.63 & 553400.6 & \textcolor{red}{\em{553400.6}} & 0.02\\

\cmidrule{1-11}

 & 18pr439-3x6 & 1475993.8 & \textcolor{black}{1498114.4} & 2.19 & 1483618.3 & \textcolor{black}{1488675.8} & 57.18 & 1471788.7 & \textcolor{red}{\em{1472798.2}} & 0.07\\
 
 & 20pr439-4x5 & 1993816.3 & \textcolor{black}{2008663.3} & 1.87 & 1993350.8 & \textcolor{black}{2000099.7} & 50.65 & 1978001.3 & \textcolor{red}{\em{1978001.3}} & 0.07\\
 
 & 25a280-5x5 & 42016.8 & \textcolor{black}{42150.6} & 0.24 & 42123.1 & \textcolor{black}{42388.8} & 13.10 & 41690.3 & \textcolor{red}{\em{41690.3}} & 0.07\\
 
 & 25gil262-5x5 & 30825.3 & \textcolor{black}{31079.8} & 0.20 & 31116.2 & \textcolor{black}{31372.9} & 11.60 & 30649.5 & \textcolor{red}{\em{30649.5}} & 0.07\\
 
 & 25pcb442-5x5 & 748160.1 & \textcolor{black}{752959.6} & 0.59 & 752573.9 & \textcolor{black}{762391.2} & 28.68 & 740883.3 & \textcolor{red}{\em{740883.3}} & 0.07\\
  \addlinespace
 & 36pcb442-6x6 & 866373.4 & \textcolor{black}{869007.5} & 0.53 & 887178.2 & \textcolor{black}{901697.8} & 30.92 & 860978.2 & \textcolor{red}{\em{860993.5}} & 0.10\\
 
 & 42a280-6x7 & 44080.5 & \textcolor{black}{44159.4} & 0.20 & 45030.0 & \textcolor{black}{46000.9} & 12.95 & 43896.8 & \textcolor{red}{\em{43896.8}} & 0.10\\
 
 & 49lin318-7x7 & 572106.1 & \textcolor{black}{574040.2} & 0.28 & 609127.8 & \textcolor{black}{633218.4} & 15.55 & 569746.3 & \textcolor{red}{\em{569755.3}} & 0.12\\
 
 & 9a280-3x3 & 29257.3 & \textcolor{black}{29497.3} & 0.36 & 29045.1 & \textcolor{black}{29105} & 14.55 & 28947.5 & \textcolor{red}{\em{28947.5}} & 0.02\\
 
 & 9gil262-3x3 & 21248.9 & \textcolor{black}{21453.8} & 0.32 & 20937.8 & \textcolor{black}{20993.1} & 14.27 & 20935.9 & \textcolor{red}{\em{20935.9}} & 0.02\\
  \addlinespace
 & 9lin318-3x3 & 720268.4 & \textcolor{black}{725327.5} & 0.47 & 718479.4 & \textcolor{black}{719450.3} & 20.27 & 716850.2 & \textcolor{red}{\em{716850.2}} & 0.02\\
 
 & 9pcb442-3x3 & 763118.7 & \textcolor{black}{769413.2} & 1.05 & 760484.4 & \textcolor{black}{761269.1} & 41.25 & 760238.3 & \textcolor{red}{\em{760238.3}} & 0.02\\
 
\multirow{-13}{*}{\centering \begin{sideways} \textbf{Large instances} \end{sideways}} 

& 9pr439-3x3 & 1824280.6 & \textcolor{black}{1857520.1} & 2.86 & 1803288.2 & \textcolor{red}{\em{1809146.2}} & 56.98 & 1800753.9 & \textcolor{black}{1830999.9} & 0.47\\
\bottomrule
\end{longtable}

\end{landscape}
\begin{table}[htbp]
  \centering
  \caption{The different results obtained by \gls{k-mfea} when running with the number of parents in range 2 to 10 }
    \begin{tabular}{c l r r r r}
    \toprule
    \multicolumn{2}{c}{\multirow{2}[4]{*}{\textbf{Instances}}} & 
    \multicolumn{4}{c}{\textbf{The number of parents}} \\
    
	\cmidrule{3-6}    
	\multicolumn{2}{c}{} & 
	\multicolumn{1}{c}{2} & 
	\multicolumn{1}{c}{3} & 
	\multicolumn{1}{c}{4} & 
	\multicolumn{1}{c}{5} \\
    \midrule
    \multirow{5}[10]{*}{\textbf{Type 1 large}} & 
    
    10pr439 & 1904718.9 & \textcolor[rgb]{ 1,  0,  0}{1904690.2} & 1904690.2 & 1904690.2 \\

	& 25pr439 & 1511197.2 & \textcolor[rgb]{ 1,  0,  0}{1511168.9} & 1511168.9 & 1511168.9 \\
           & 50lin318 & 688748.63 & \textcolor[rgb]{ 1,  0,  0}{688724.60} & 688724.60 & 688724.60 \\
           & 50pcb442 & 910487.55 & \textcolor[rgb]{ 1,  0,  0}{910480.10} & 910478.70 & 910478.70 \\
           & 50pr439 & 2152987.86 & \textcolor[rgb]{ 1,  0,  0}{2152986.60} & 2152986.60 & 2152986.60 \\
    \midrule
    \textbf{Type 3} & 6i500 & 37516.12 & \textcolor[rgb]{ 1,  0,  0}{37,510} & 37,510 & 37,510 \\
           
    \midrule
    \multirow{2}[4]{*}{\textbf{Type 5 Large}} & 20i500-307 & 200343 & \textcolor[rgb]{ 1,  0,  0}{200328.7} & 200328.7 & 200328.7 \\
           & 5i500-304 & 183976.6 & \textcolor[rgb]{ 1,  0,  0}{182446} & \textcolor[rgb]{ 1,  0,  0}{182434.5} & \textcolor[rgb]{ 1,  0,  0}{182403} \\
    \midrule
    \multirow{4}[8]{*}{\textbf{Type 6 Large}} & \multicolumn{1}{p{6.11em}}{18pr439-3x6} & 1472798.2 & \textcolor[rgb]{ 1,  0,  0}{1471788.7} & 1471788.7 & 1471788.7 \\
    
           & \multicolumn{1}{l}{9pr439-3x3} & \multicolumn{1}{r}{1830999.9} & \multicolumn{1}{r}{\textcolor[rgb]{ 1,  0,  0}{1810272.6}} & \textcolor[rgb]{ 1,  0,  0}{1808494.3} & \textcolor[rgb]{ 1,  0,  0}{1800753.9} \\
           & \multicolumn{1}{p{6.11em}}{36pcb442-6x6} & 860993.5 & \textcolor[rgb]{ 1,  0,  0}{860978.2} & 860978.2 & 860978.2 \\
           & \multicolumn{1}{p{6.11em}}{49lin318-7x7} & 569755.3 & \textcolor[rgb]{ 1,  0,  0}{569746.3} & 569746.3 & 569746.3 \\
   
    \bottomrule
    \end{tabular}%
  \label{tab:Results_Parents}%
\end{table}%

\end{document}